\documentclass[nohyperref,dvipsnames]{article}
\usepackage[T1]{fontenc}
\usepackage{microtype}
\usepackage{graphicx}
\usepackage{subcaption}
\usepackage{booktabs} 

\usepackage{hyperref}

\usepackage[accepted]{icml2022}

\usepackage{amsmath}
\usepackage{amssymb}
\usepackage{mathtools}
\usepackage{amsthm}

\usepackage{tikz}
\newcommand{\tikzcircle}[2][red,fill=red]{\tikz[baseline=-0.5ex]\draw[#1,radius=#2] (0,0) circle ;}%

\usepackage[capitalize]{cleveref}

\theoremstyle{plain}

\theoremstyle{definition}

\theoremstyle{remark}

\usepackage[textsize=tiny]{todonotes}

\icmltitlerunning{Tackling covariate shift with node-based BNNs}

\def\f{\mathbf{f}}
\def\hf{\hat{\f}}

\def\x{\mathbf{x}}

\def\z{\mathbf{z}}

\def\g{\mathbf{g}}
\def\htheta{\hat{\theta}}
\def\1{\mathbf{1}}
\def\0{\mathbf{0}}

\def\H{\mathbb{H}}

\def\L{\mathcal{L}}
\def\y{\mathbf{y}}

\def\W{\mathbf{W}}
\def\s{\mathbf{s}}

\def\Z{\mathcal{Z}}
\def\J{\mathbf{J}}

\def\bmu{\boldsymbol\mu}
\def\bs{\boldsymbol\sigma}

\def\0{\mathbf{0}}
\def\N{\mathcal{N}}
\def\D{\mathcal{D}}
\def\b{\mathbf{b}}
\def\h{\mathbf{h}}
\def\E{\mathbb{E}}

\def\diag{\mathrm{diag}}

\def\kl{\mathrm{KL}}

\DeclareMathOperator*{\argmax}{arg\,max}
\DeclareMathOperator*{\argmin}{arg\,min}

\begin{document}

\twocolumn[
\icmltitle{Tackling covariate shift with node-based Bayesian neural networks}



\icmlsetsymbol{equal}{*}

\begin{icmlauthorlist}
\icmlauthor{Trung Trinh}{aalto}
\icmlauthor{Markus Heinonen}{aalto}
\icmlauthor{Luigi Acerbi}{helsinki}
\icmlauthor{Samuel Kaski}{aalto,manchester}
\end{icmlauthorlist}

\icmlaffiliation{aalto}{Department of Computer Science, Aalto University, Finland}
\icmlaffiliation{helsinki}{Department of Computer Science, University of Helsinki, Finland}
\icmlaffiliation{manchester}{Department of Computer Science, University of Manchester, UK}

\icmlcorrespondingauthor{Trung Trinh}{trung.trinh@aalto.fi}

\icmlkeywords{Machine Learning, ICML}

\vskip 0.3in
]



\printAffiliationsAndNotice{}  

\begin{abstract}
Bayesian neural networks (BNNs) promise improved generalization under covariate shift by providing principled probabilistic representations of epistemic uncertainty. 
However, weight-based BNNs often struggle with high computational complexity of large-scale architectures and datasets.
Node-based BNNs have recently been introduced as scalable alternatives, which induce epistemic uncertainty by multiplying each hidden node with latent random variables, while learning a point-estimate of the weights.
%
In this paper, we interpret these latent noise variables as implicit representations of simple and domain-agnostic data perturbations during training, producing BNNs that perform well under covariate shift due to input corruptions.
We observe that the diversity of the implicit corruptions depends on the entropy of the latent variables, and propose a straightforward approach to increase the entropy of these variables during training.
We evaluate the method on out-of-distribution image classification benchmarks, and show improved uncertainty estimation of node-based BNNs under covariate shift due to input perturbations. %
As a side effect, the method also provides robustness against noisy training labels.
\end{abstract}

\section{Introduction}
Bayesian neural networks (BNNs) induce epistemic uncertainty over predictions by placing a distribution over the weights \citep{mackay1992,mackay1995,hinton1993,neal_bayesian_1996}.
However, it is challenging to infer the weight posterior due to the high dimensionality and multi-modality of this distribution \citep{wenzel2020good,izmailov2021bayesian}.
Alternative BNN methods have been introduced to avoid the complexity of weight-space inference, which include combining multiple maximum-a-posteriori (MAP) solutions \citep{lakshminarayanan2017simple}, performing inference in the function-space \citep{sun2019functional}, or performing inference in a lower dimensional latent space \citep{karaletsos2018probabilistic,pradier2018projected,izmailov20a,dusenberry20a}.

A recent approach to simplify BNNs is \emph{node stochasticity}, which assigns latent noise variables to hidden nodes of the network \citep{variational_dropout,gal2016dropout,karaletsos2018probabilistic,karaletsos2020hierarchical,dusenberry20a,nguyen2021structured}.
By restricting inference to the node-based latent variables, node stochasticity greatly reduces the dimension of the posterior, as the number of nodes is orders of magnitude smaller than the number of weights in a neural network \citep{dusenberry20a}.
Within this framework, multiplying each hidden node with its own random variable has been shown to produce great predictive performance, while having dramatically smaller computational complexity compared to weight-space BNNs \citep{gal2016dropout,variational_dropout,dusenberry20a,nguyen2021structured}. 

In this paper, we focus on \emph{node-based BNNs}, which represent epistemic uncertainty by inferring the posterior distribution of the multiplicative latent node variables while learning a point-estimate of the weight posterior \cite{dusenberry20a,trinh2020scalable}.
We show that node stochasticity simulates a set of implicit corruptions in the data space during training, and by learning in the presence of such corruptions, node-based BNNs achieve natural robustness against some real-world input corruptions.
This is an important property because one of the key promises of BNNs is robustness under \emph{covariate shift} \citep{ovadia2019can,izmailov2021bayesian}, defined as a change in the distribution of input features at test time with respect to that of the training data.
Based on our findings, we derive an entropy regularization approach to improve out-of-distribution generalization for node-based BNNs.
In summary, our contributions are:
\begin{enumerate}
    \item We demonstrate that node stochasticity simulates data-space corruptions during training. We show that the diversity of these corruptions corresponds to the entropy of the latent node variables, and training on more diverse generated corruptions produce node-based BNNs that are robust against a wider range of corruptions.
    \item We derive an entropy-regularized variational inference formulation for node-based BNNs.
    \item We demonstrate excellent empirical results in predictive uncertainty estimation under covariate shift due to corruptions compared to strong baselines on large-scale image classification tasks.
    \item We show that, as a side effect, our approach  provides robust learning in the presence of noisy training labels.
\end{enumerate}
Our code is available at \url{https://github.com/AaltoPML/node-BNN-covariate-shift}.
\section{Background}

\paragraph{Neural networks.} We define a standard neural network $\f(\x)$ with $L$ layers for an input $\x$ as follows:
\begin{align}
    \f^0(\x) &= \x \label{eq:f0}\\
    \h^\ell(\x) &= \W^\ell \f^{\ell-1}(\x) + \b^\ell\\
    \f^\ell(\x) &= \sigma^\ell \big( \h^\ell(\x) \big), \qquad \forall \ell=1,\dots,L \label{eq:f_ell}\\
    \f(\x) &= \f^L(\x), \label{eq:fL}
\end{align}
where the parameters $\theta=\{\W^\ell, \b^\ell\}_{\ell=1}^L$ consist of the weights and biases, and the $\{\sigma^\ell\}_{\ell=1}^L$ are the activation functions. 
For the $\ell$-th layer, $\h^\ell$ and $\f^\ell$ are the pre- and post-activations, respectively.

\paragraph{Node-based Bayesian neural networks.} Probabilistic neural networks constructed using node stochasticity have been studied by \citet{gal2016dropout,variational_dropout,louizos2017multiplicative, karaletsos2018probabilistic, karaletsos2020hierarchical, dusenberry20a, trinh2020scalable, nguyen2021structured}. 
We focus on inducing node stochasticity by multiplying each hidden node with its own random latent variables, and follow the framework of \citet{dusenberry20a} for optimization. A node-based BNN $\f_\Z(\x)$ is defined as:
\begin{align}
    \f^0_\Z(\x) &= \x \label{eq:f0_z} \\
    \h^\ell_\Z(\x) &=  (\W^\ell (\f^{\ell-1}_\Z (\x) \circ \z^\ell ) + \b^\ell) \circ \s^\ell \label{eq:h_ell_z} \\
    \f_\Z^\ell(\x) &= \sigma^\ell \big ( \h_\Z^\ell(\x) \big ), \qquad \forall \ell=1,\dots,L \label{eq:f_ell_z} \\
    \f_\Z(\x) &= \f^L_\Z(\x), \label{eq:fL_z}
\end{align}
where $\z^\ell$ and $\s^\ell$ are the multiplicative latent random variables of the incoming and outgoing signal of the nodes of the $\ell$-th layer, and $\circ$ denotes the Hadamard (element-wise) product. 
We collect all latent variables to $\Z=\{\z^\ell, \s^\ell\}_{\ell=1}^L$.\footnote{In this paper, we use a slightly more general definition of node-based BNN with two noise variables per node, and compare it with single-variable variants in Section \ref{sec:experiments}.}

To learn the network parameters, we follow \citet{dusenberry20a} and perform variational inference \citep{blei2017variational} over the weight parameters $\theta$ and latent node variables $\Z$. We begin by defining a prior $p(\theta, \Z) = p(\theta)p(\Z)$. We set a variational posterior approximation $q_{\hat{\theta},\phi}(\theta,\Z)=q_{\hat{\theta}}(\theta)q_\phi(\Z)$, where $q_{\hat{\theta}}(\theta)=\delta(\theta-\hat{\theta})$ is a Dirac delta distribution and $q_\phi(\Z)$ is a Gaussian or a mixture of Gaussians distribution. 
We infer the posterior by minimizing the Kullback-Leibler (KL) divergence between variational approximation $q$ and true posterior $p(\theta, \Z|\D)$.
This is equivalent to maximizing the evidence lower bound (ELBO):
\begin{multline}
    \L(\htheta, \phi) = \E_{q_\phi(\Z)} \Big[ \log p(\D | \htheta,\Z) \Big]
	\\ - \kl\Big[q_\phi(\Z) \, || \, p(\Z)\Big]  + \log p(\htheta). \label{eq:elbo}
\end{multline}
In essence, we find a MAP solution for the more numerous weights $\theta$, while inferring the posterior distribution of the latent variables $\Z$. We refer the reader to \cref{appendix:orig_elbo} for detailed derivations.

\paragraph{Neural networks under covariate shift.} 

\begin{figure}[t]
    \centering
    \includegraphics[width=.80\columnwidth]{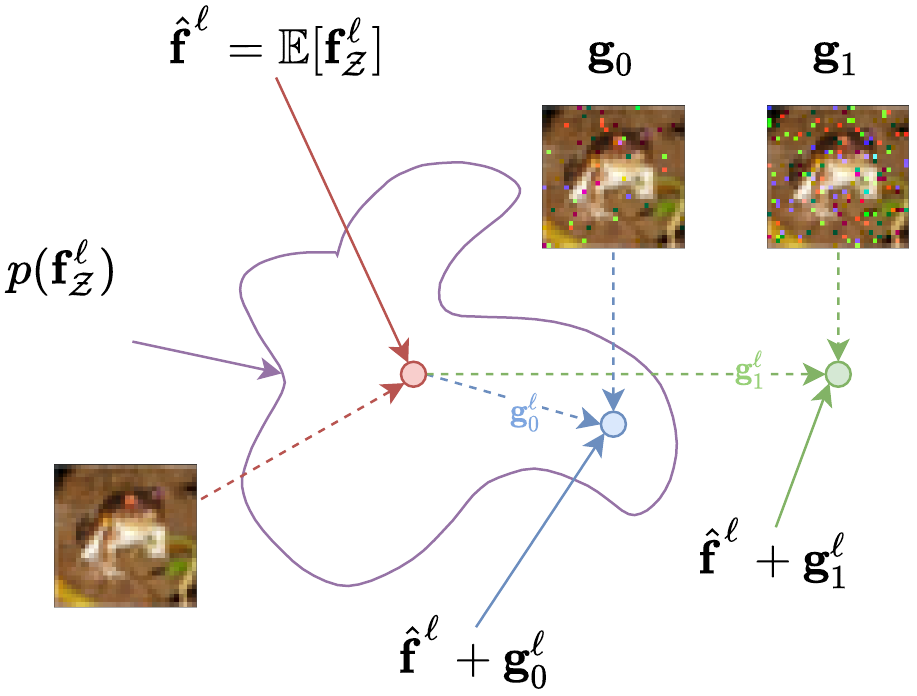}
    \caption{A sketch depicting the connection between the output distribution at the $\ell$-th layer induced by node stochasticity (\textcolor{rgb,255:red,150;green,115;blue,166}{purple}) centered on the average output (\tikzcircle[fill={rgb,255:red, 248; green, 206; blue,204},draw={rgb,255: red,184; green, 84; blue, 80}]{2.5pt}), and the output shifts generated by input corruptions (\tikzcircle[fill={rgb,255: red,218; green,232; blue,252},draw={rgb,255:red,108;green,142;blue,191}]{2.5pt},\tikzcircle[fill={rgb,255:red,213;blue,232;green,212},draw={rgb,255:red,130;green,179;blue,102}]{2.5pt}). We expect good performance under mild corruption $\g_0$, as the resulting shift remains inside the high-density region of $p(\f^\ell_\Z)$, and worse results under severe corruption $\g_1$.}
    \label{fig:entropy_shift_relation_conceptual}
\end{figure}

In this paper, we focus on covariate shift from input corruptions, following the setting of \citet{hendrycks2019robustness}. %
To simulate covariate shift, one can take an input $\x$ assumed to come from the same distribution as the training samples and apply an input corruption $\g^0$ to form a shifted version $\x^c$ of $\x$:
\begin{equation}
    \x^c = \x + \g^0(\x).
\end{equation}
For instance, $\x$ could be an image and $\g^0$ can represent the shot noise corruption \citep{hendrycks2019robustness}.
The input corruption $\g^0(\x)$ creates a shift in the output of each layer $\g^\ell(\x)$ (see \cref{fig:entropy_shift_relation_conceptual}). We can approximate these shifts by first-order Taylor expansion (see \cref{appendix:activation_shifts} for full derivation),
\begin{align}\label{eq:shift_ell}
    \overbrace{\g^\ell(\x)}^{\text{shift}} &= \overbrace{\f^\ell(\x^c)}^{\text{corrupted output}} - \overbrace{\f^\ell(\x)}^{\text{clean output}} \\
    &\approx \J_\sigma \big[ \h^\ell(\x) \big] \big(\W^{\ell}\g^{\ell-1}(\x)\big), \label{eq:shift_ell_approx}
\end{align}
where $\J_{\sigma^\ell} = \partial \sigma^{\ell}/\partial \h^\ell$ denotes the (diagonal) Jacobian of the activation $\sigma^\ell$ with respect to $\h^\ell$. 
While $\g^0$ causes activation shifts in every layer of the network, we focus on the shift in the final output layer $\g^L$.
The approximation in \cref{eq:shift_ell_approx} shows that this shift depends on the input $\x$, the network's architecture (e.g., choice of activation functions) and parameters $\theta$.
We measure the robustness of a network with respect to a corruption $\g^0(\cdot)$ on the dataset $\D=\{\x_n, \y_n\}_{n=1}^N$ by the induced \emph{mean square shift},
\begin{align}
    \mathrm{MSS}_g &= \frac{1}{N} \sum_{n=1}^N ||\g^L(\x_n)||^2_2, \label{eq:AS_g}
\end{align}
where $\mathrm{MSS}_g$ is the average shift on the data.
Ideally, we want $\mathrm{MSS}_g$ to be small for the network to still provide nearly correct predictions given corrupted inputs.
When the training data and the architecture are fixed, $\mathrm{MSS}_g$ depends on the parameters $\theta$.
A direct approach to find $\theta$ minimizing $\mathrm{MSS}_g$ is to apply the input corruption $\g^0$ to each input $\x_n$ during training to teach the network to output the correct label $\y_n$ given $\g^0(\x_n)$.
However, this approach is not domain-agnostic and requires defining a list of corruptions beforehand.
In the next sections, we discuss the usage of multiplicative latent node variables as an implicit way to simulate covariate shifts during training.

\section{Characterizing implicit corruptions}\label{sec:implicit_corruption}


In this section, we demonstrate that multiplicative node variables correspond to implicit input corruptions. We show how to extract and visualize these new corruptions.

\subsection{Relating input corruptions and multiplicative nodes}\label{sec:input_corruption_and_latent_variables}

The node-based BNN of \crefrange{eq:f0_z}{eq:fL_z} induces the \emph{predictive posterior} $p(\f^\ell_\Z(\x))$ over the $\ell$-th layer outputs by marginalizing over the variational latent \emph{parameter posterior} $q(\Z_{\leq \ell})$.
Optimization of the variational objective in \cref{eq:elbo} enforces the model to achieve low loss on the training data despite each layer output being corrupted by noise from $q(\Z)$, represented by the expected log likelihood term of the ELBO. 
Let $\hf^\ell(\x)$ denote the mean predictive posterior,
\begin{align}
    \hf^\ell(\x) &= \E_{q(\Z)} \big[\f_\Z^\ell(\x)\big], \quad \forall \ell=1,\ldots,L,
\end{align}
and where we denote the final output $\hf(\x) = \hf^L(\x)$.
If the shifted output $\hf^\ell(\x+\g^0(\x)) = \hf^\ell(\x) + \g^\ell(\x)$ caused by corrupting a training sample $\x$ using $\g^0$ lies within the predictive distribution of $\f^\ell_\Z(\x)$ (blue dot in \cref{fig:entropy_shift_relation_conceptual}), then the model can map this corrupted version of $\x$ to its correct label.
This implies robustness against the space of implicit corruptions generated by $q(\Z)$, which indirectly leads to robustness against real corruptions.
However, standard variational inference will converge to a posterior whose entropy is calibrated for the variability in the training data, but does not necessarily account for corruptions caused by covariate shifts. Thus, the posterior might cover the corruption with low severity $\g_0$ (blue dot in \cref{fig:entropy_shift_relation_conceptual}), but not the one with higher severity $\g_1$ (green dot in \cref{fig:entropy_shift_relation_conceptual}). To promote predictive distributions that are more robust to perturbations, we propose to increase the entropy of $p(\f_\Z^\ell(\x))$ by increasing the entropy of the variational posterior $q(\Z)$.

\begin{figure}[t]
    \centering
    \begin{subfigure}{.49\columnwidth}
        \includegraphics[width=\textwidth]{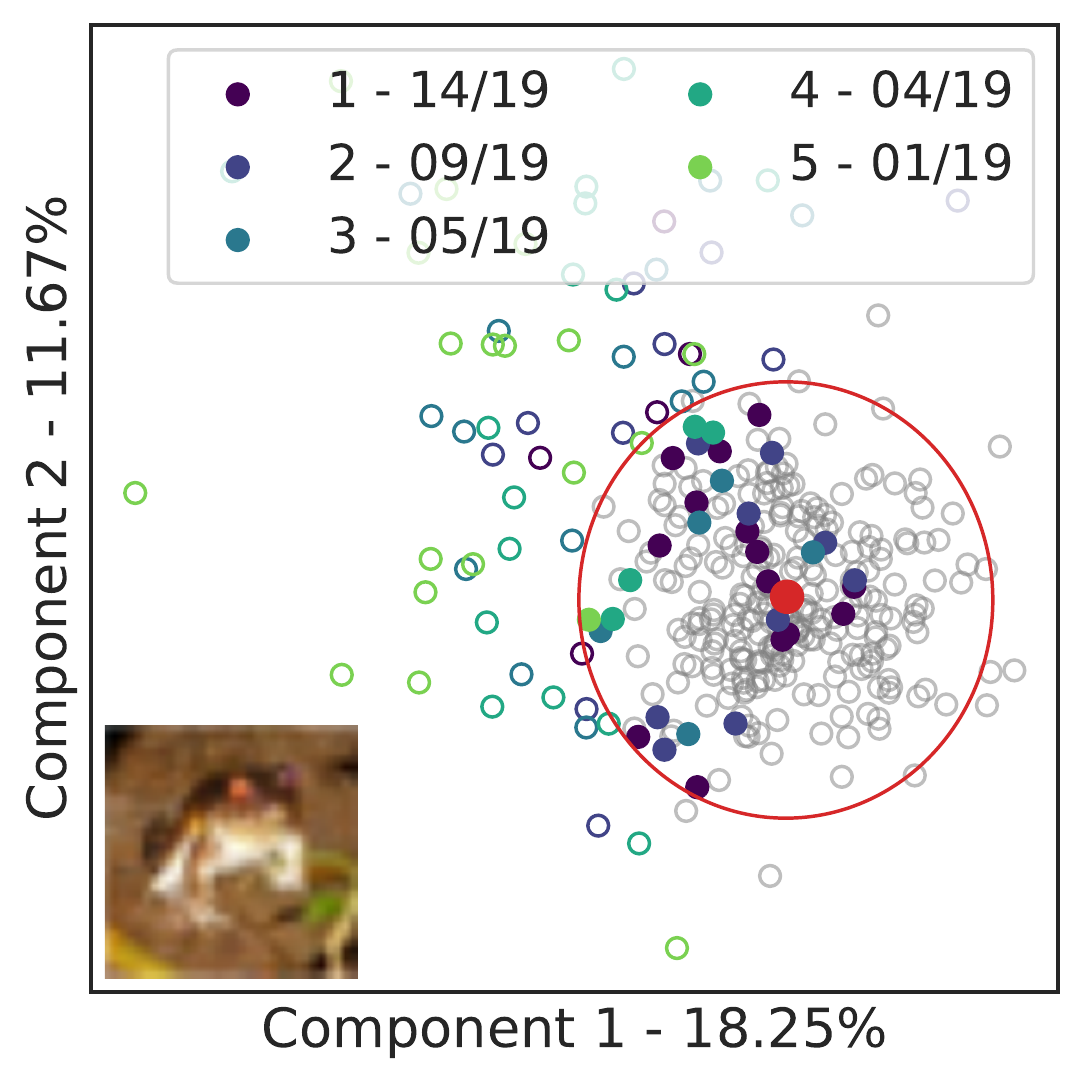}
    \end{subfigure}
    \begin{subfigure}{.49\columnwidth}
        \includegraphics[width=\textwidth]{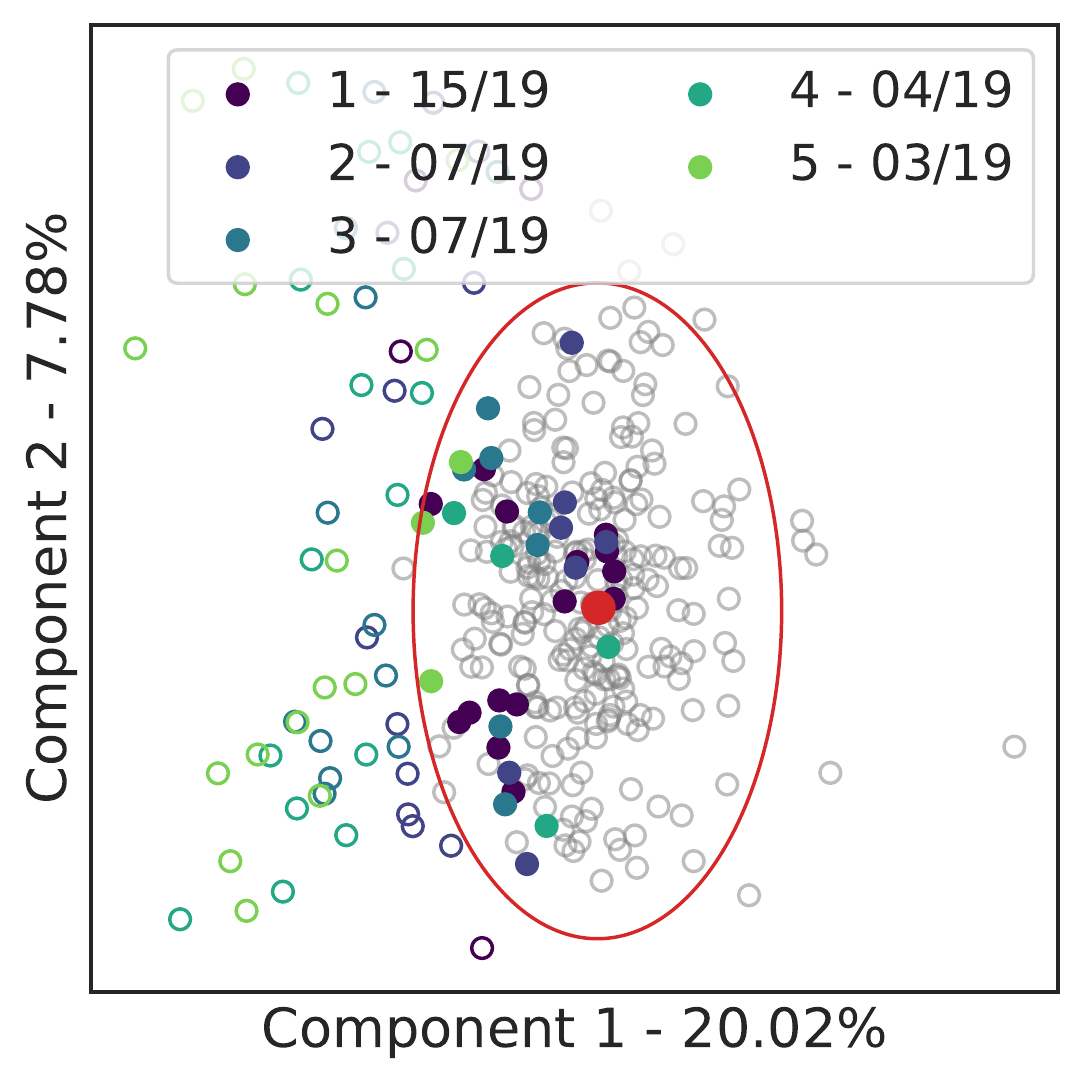}
    \end{subfigure}
    \begin{subfigure}{.49\columnwidth}
        \includegraphics[width=\textwidth]{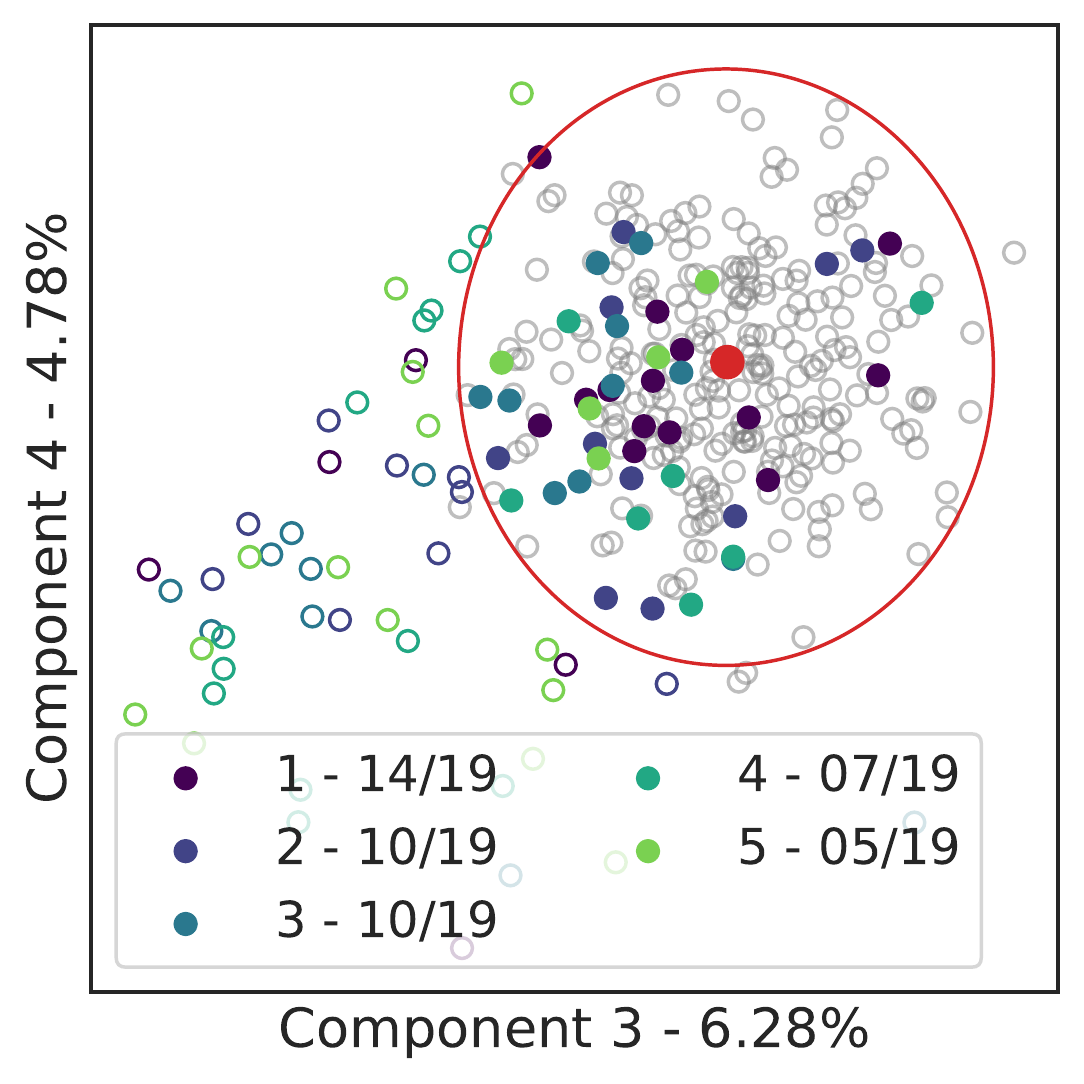}
        \caption{$\mathcal{M}_{16}$}
        \label{fig:conv8_distribution_corruption_16}
    \end{subfigure}
    \begin{subfigure}{.49\columnwidth}
        \includegraphics[width=\textwidth]{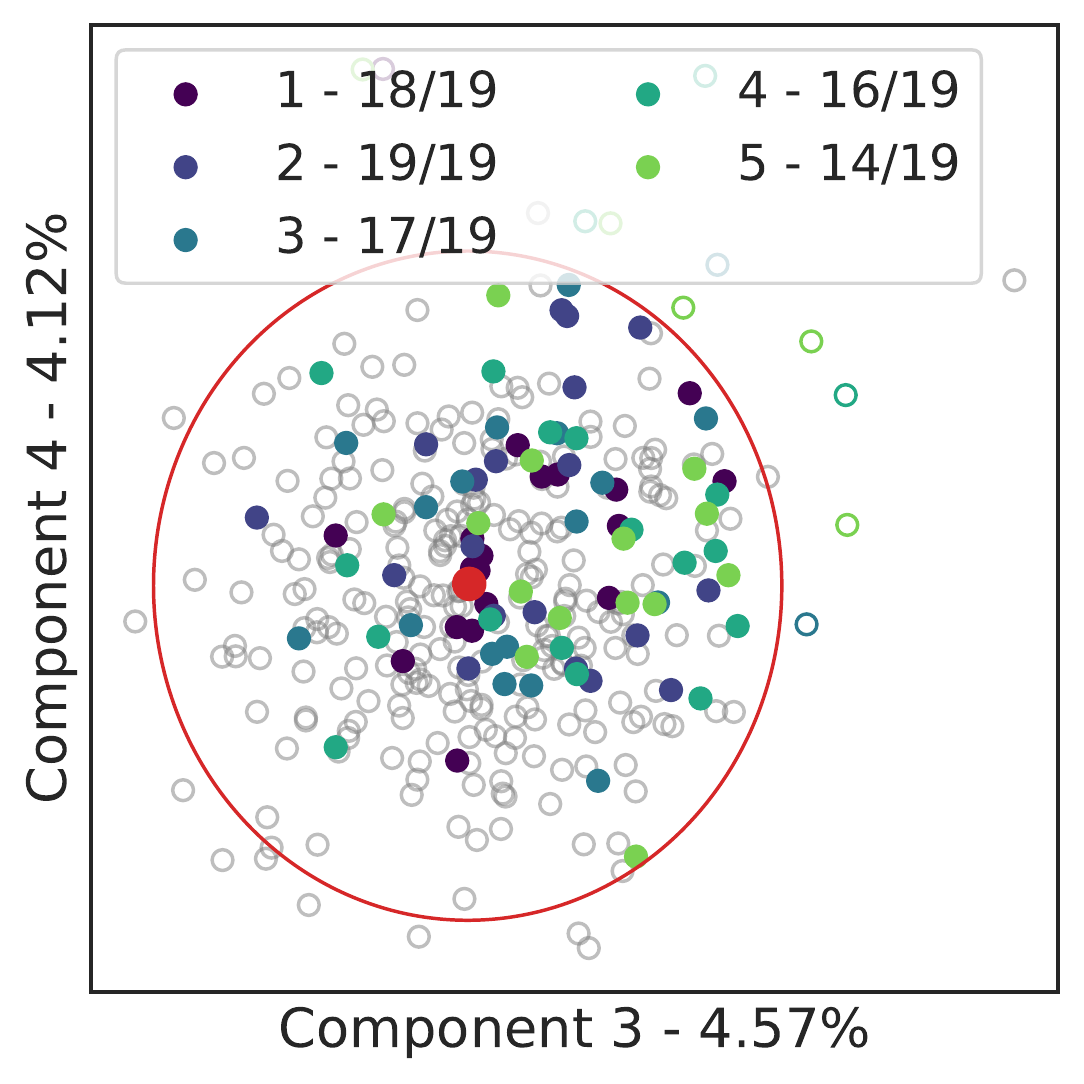}
        \caption{$\mathcal{M}_{32}$}
        \label{fig:conv8_distribution_corruption_32}
    \end{subfigure}
    \caption{PCA plots of the last layer's outputs of models \textbf{(a)} $\mathcal{M}_{16}$ and \textbf{(b)} $\mathcal{M}_{32}$ with respect to one sample from \textsc{cifar-10} (included in the top left panel). Grey circles are samples from the output distribution induced by $q(\Z)$, while the red ellipse shows their $99$ percentile. The red circle denotes the expected output $\hf^\ell(\x)=\E_{q(\Z)}[\f_\Z^\ell(\x)]$ of the test point. Other colored circles represents the expected output $\hf^\ell$ of the 19 corrupted versions of the test point under 5 levels of severity \citet{hendrycks2019robustness}. Most of the mild corruptions reside inside the predictive posterior of both models (filled color circles). By contrast, only the higher-entropy $\mathcal{M}_{32}$ model encapsulates a large fraction of the severe corruptions --
    empirically demonstrating the intuition sketched in \cref{fig:entropy_shift_relation_conceptual} and described in Section \ref{sec:input_corruption_and_latent_variables}.
    }
    \label{fig:conv8_distribution_corruption}
\end{figure}
\begin{figure}
    \centering
    \begin{subfigure}[c][3.5cm][c]{.49\columnwidth}
        \includegraphics[height=3.5cm]{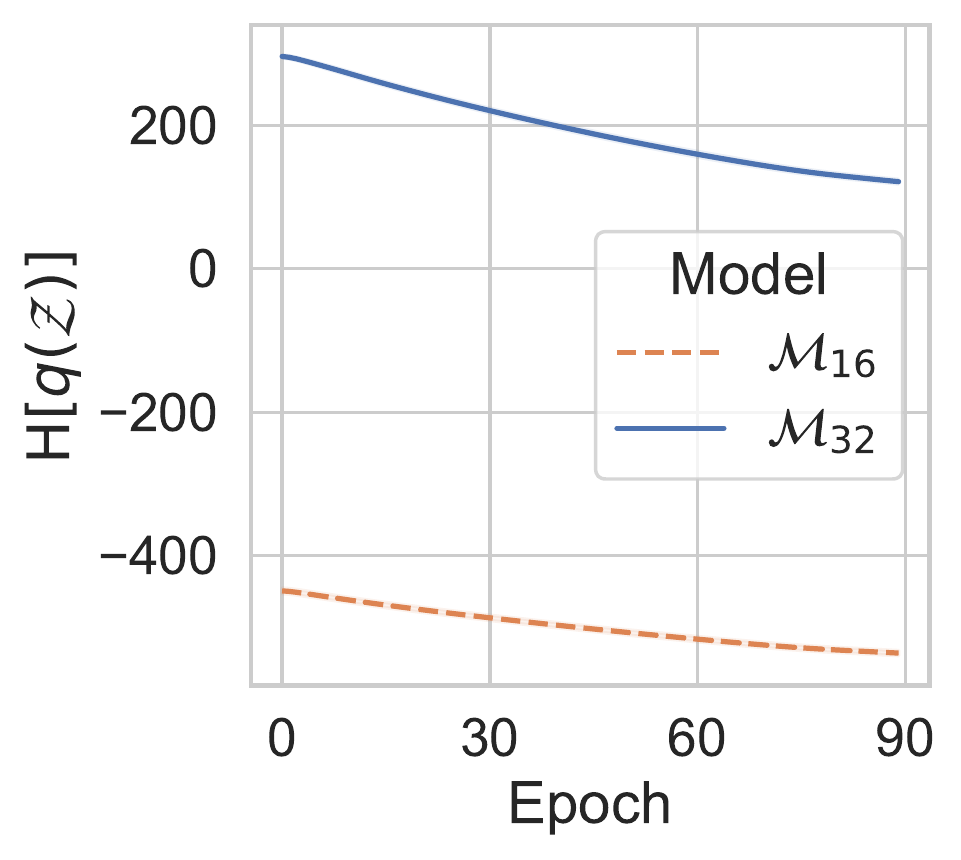}
    \end{subfigure}
    \begin{subfigure}[c][3.5cm][c]{.49\columnwidth}
        \includegraphics[height=3.5cm]{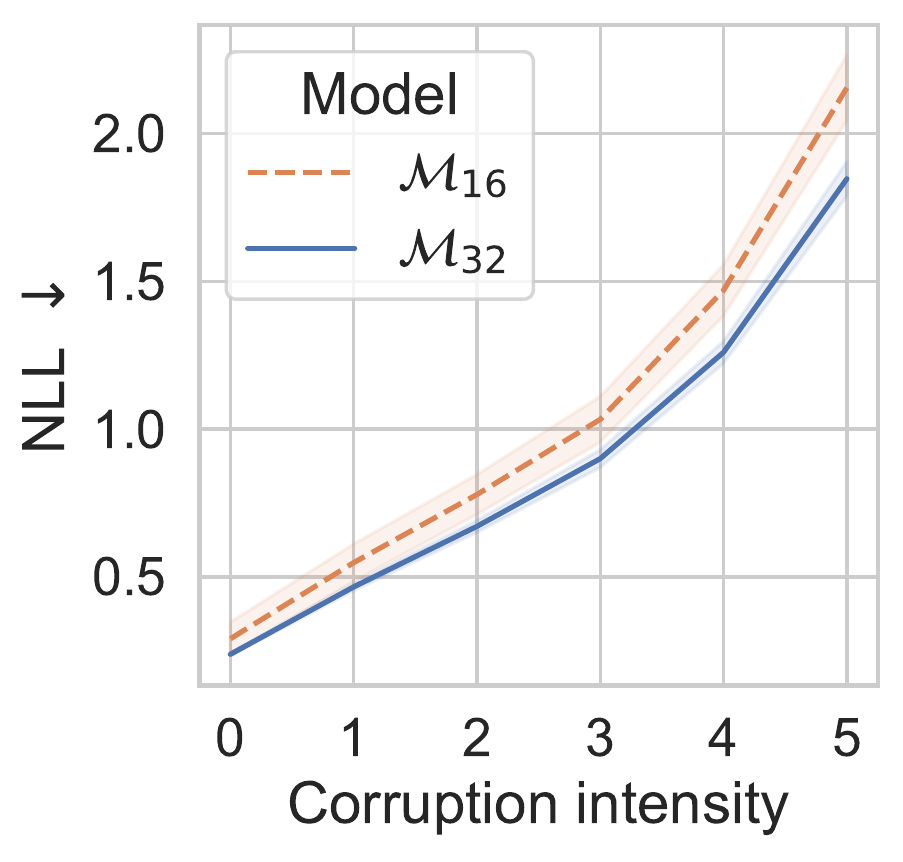}
    \end{subfigure}
    \caption{\textbf{(Left)} Example evolution of $\H[q(\Z)]$ during training, which shows that the variational entropy decreases over time. \textbf{(Right)} Performance of two models under different corruption levels (level 0 indicates no corruption). The model with higher entropy $\mathcal{M}_{32}$ performs better than the one with lower entropy $\mathcal{M}_{16}$ across all corruption levels. For each result in both plots, we report the mean and standard deviation over 25 runs. The error bars in the left plot are too small to be seen.}
    \label{fig:entropy_plot}
\end{figure}

\paragraph{Empirical demonstration.} To illustrate our intuition, we present an example with two node-based BNNs, one with high entropy and one with lower entropy.
We use the \textsc{all-cnn-c} architecture of \citet{springenberg2014striving} and \textsc{cifar10} \cite{krizhevsky2009cifar}.
We initialize the standard deviations of $q(\Z)$ for the low-entropy model using the half-normal $\N^+(0.16, 0.02)$, while we use $\N^+(0.32, 0.02)$ for the high-entropy model. 
For brevity, we refer to the former model as $\mathcal{M}_{16}$ and the latter model as $\mathcal{M}_{32}$.
In the left plot of \cref{fig:entropy_plot}, we show that, after training, $\mathcal{M}_{32}$ retains higher variational posterior entropy than $\mathcal{M}_{16}$ due to having higher initial standard deviations for $q(\Z)$.\footnote{Obtaining high-entropy models by starting with high-entropy initializations is a simple heuristic for the purpose of this example. We introduce a principled approach in Section \ref{sec:maximize_entropy}.}
We use principal component analysis (PCA) to visualize the samples from the output distribution $p(\f^\ell_\Z(\x))$ of the $\ell$-th layer with respect to one input image $\x$, as well as the output $\{\hf^\ell(\x+\g_i(\x))\}_{i=1}^{95}$ under the real image corruptions $\{\g_i\}_{i=1}^{95}$ from \citet{hendrycks2019robustness}.
There are 19 corruption types with 5 levels of severity, totalling 95 corruption functions.
\cref{fig:conv8_distribution_corruption} shows the activations of the last layer, projected into a two-dimensional subspace with PCA for visualization.
From this figure, we can see that there is more overlap between samples from $p(\f_\Z^\ell(\x))$ and the shifted outputs $\{\hf^\ell(\x+\g_i(\x))\}_{i=1}^{95}$ for $\mathcal{M}_{32}$ in \cref{fig:conv8_distribution_corruption_32} than for $\mathcal{M}_{16}$ in \cref{fig:conv8_distribution_corruption_16}.
This indicates that during training the posterior of $\mathcal{M}_{32}$ is able to simulate a larger number of implicit corruptions bearing resemblance to the real-world corruptions than the posterior of $\mathcal{M}_{16}$, leading to better negative log-likelihood (NLL) accross all level of corruptions as well as on the clean test set in \cref{fig:entropy_plot}.
This example supports our intuition that increasing the entropy of the latent variables $\Z$ allows them to simulate more diverse implicit corruptions, thereby boosting the model's robustness against a wider range of input corruptions.




\begin{figure}[t]
    \centering
    \begin{subfigure}{0.32\columnwidth}
    \includegraphics[width=\textwidth]{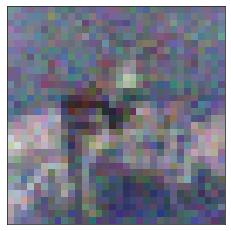}
    \end{subfigure}
    \hfill
    \begin{subfigure}{0.32\columnwidth}
    \includegraphics[width=\textwidth]{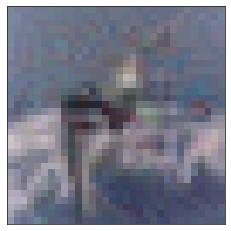}
    \end{subfigure}
    \hfill
    \begin{subfigure}{0.32\columnwidth}
    \includegraphics[width=\textwidth]{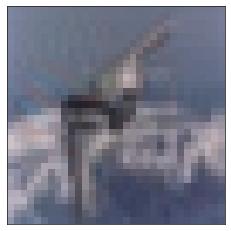}
    \end{subfigure}
    \begin{subfigure}{0.32\columnwidth}
    \includegraphics[width=\textwidth]{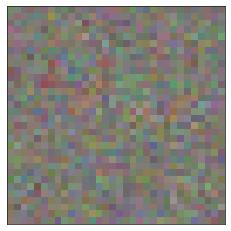}
    \caption{$\lambda = 0.03$}
    \label{fig:mask_005}
    \end{subfigure}
    \hfill
    \begin{subfigure}{0.32\columnwidth}
    \includegraphics[width=\textwidth]{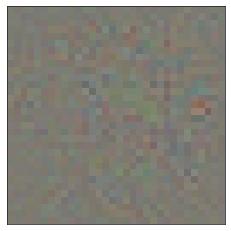}
    \caption{$\lambda = 0.10$}
    \label{fig:mask_01}
    \end{subfigure}
    \hfill
    \begin{subfigure}{0.32\columnwidth}
    \includegraphics[width=\textwidth]{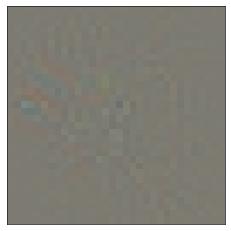}
    \caption{$\lambda = 0.30$}
    \label{fig:mask_05}
    \end{subfigure}
    \caption{Implicit corruptions generated from model $\mathcal{M}_{32}$ with respect to one image by minimizing the loss in \cref{eq:xc_loss} under varying $\lambda$. Top row are the resulting images from the corruptions below. We can see that $\lambda$ controls the severity of the generated corruptions.}
    \label{fig:implict_corruptions}
\end{figure}
\paragraph{Why latent variables at every layer?} In principle, we could have introduced latent variables only to the first layer of the network, as the shift simulated in the first layer will propagate to subsequent layers.
However, modern NNs contain asymmetric activation functions such as ReLU or Softplus, which can attenuate the signal of the shift in the later layers.
Thus, the latent variables in every layer (after the first one) maintain the strength of the shift throughout the network during the forward pass.
Moreover, by using latent variables at every layer -- as opposed to only the first layer -- we can simulate a more diverse set of input corruptions, since we can map each sample $\Z$ from $q(\Z)$ to an input corruption as shown in the following section.


\subsection{Visualizing the implicit corruptions}

\begin{figure}[t]
    \centering
    \begin{subfigure}[c][3.5cm][c]{.49\columnwidth}
        \centering
        \includegraphics[height=3.5cm]{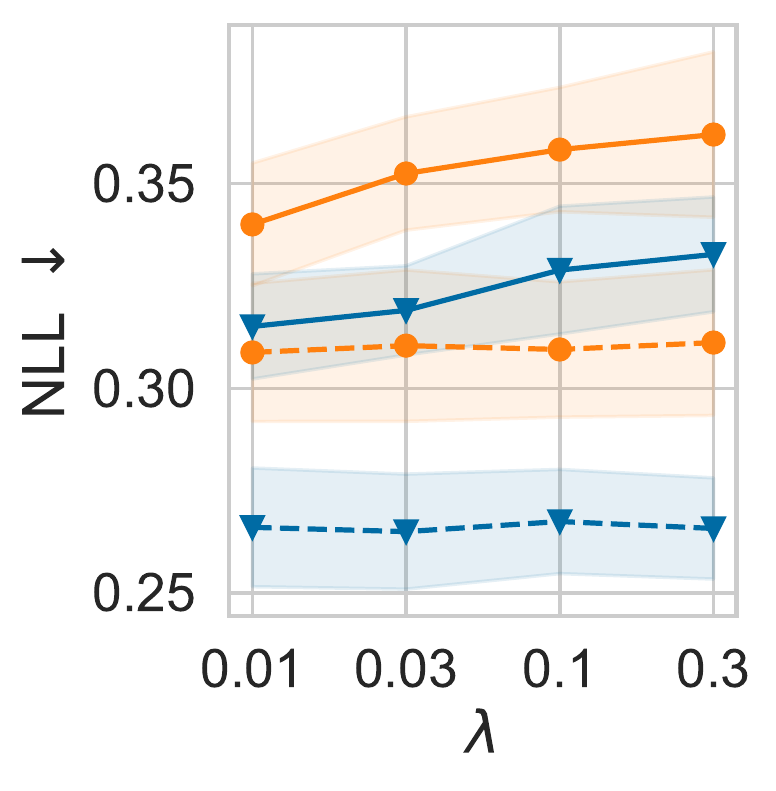}
    \end{subfigure}
    \hfill
    \begin{subfigure}[c][3.5cm][c]{.49\columnwidth}
        \centering
        \includegraphics[height=3.5cm]{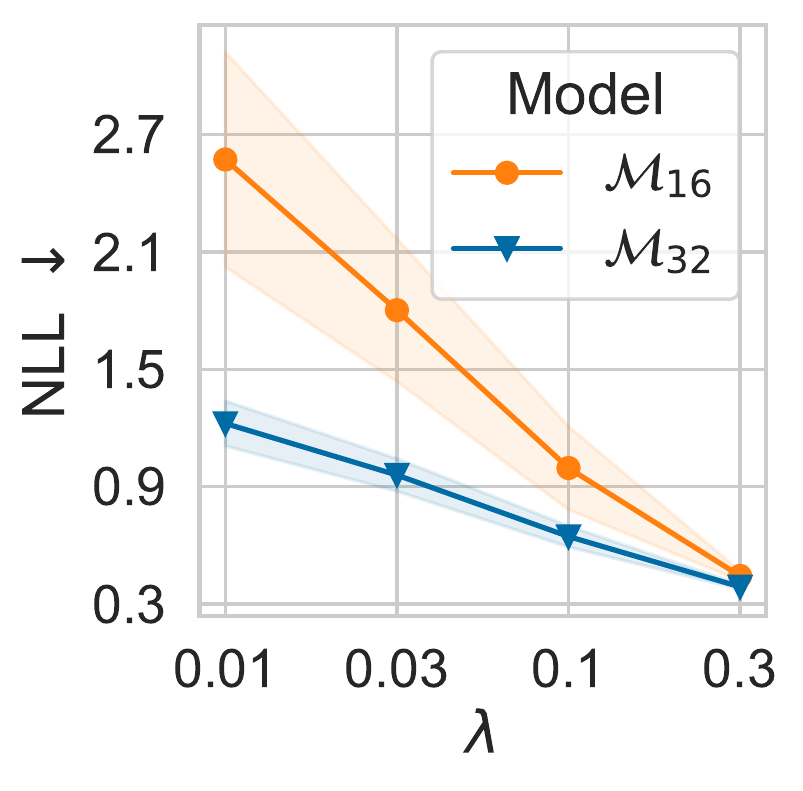}
    \end{subfigure}
    \caption{Negative log-likelihood (NLL) on 1024 test images of \textsc{cifar-10} corrupted by the implicit corruptions generated by $\mathcal{M}_{16}$ and $\mathcal{M}_{32}$, whose intensities are controlled by $\lambda$ in \cref{eq:xc_loss}. For each result, we report the mean and standard deviation over 10 runs. \textbf{(Left)} Each model is tested on the corruptions that it generated. Dashed lines are results on the clean images for reference. Each model is resistant to its own corruptions, as evidenced by the slight decrease in performance under different $\lambda$. \textbf{(Right)} Each model is tested on the corruptions produced by the other. The model with higher entropy $\mathcal{M}_{32}$ is more robust against the corruptions of the one with lower entropy $\mathcal{M}_{16}$ than the reverse, which further supports the notion that higher entropy provides better robustness against input corruptions.}
    \label{fig:cross_testing_implicit_corruption}
\end{figure}


Next, we show how to find the explicit image corruptions that correspond to the stochasticity of the predictive posterior. Let $\Z$ be a sample drawn from $q(\Z)$. If we assume that $\Z$ corresponds to an input corruption $\g(\x)$:
\begin{equation}
    \f_{\Z} (\x) = \hf(\x + \g(\x))
\end{equation}
then we can approximately solve for $\g(\x) = \x^c - \x$ by finding $\x^c$ that minimizes
\begin{equation}\label{eq:xc_loss}
    \L(\x^c) = \frac{1}{2}\left|\left|\f_{\Z}(\x)-\hf(\x^c)\right|\right|^2_2 + \frac{\lambda}{2}\big|\big| \g(\x) \big|\big|_2^2
\end{equation}
using gradient descent. The second term with a coefficent $\lambda \geq 0$ regularizes the norm of $\g(\x)$. This approach is similar to the method of finding adversarial examples of \citet{goodfellow2014explaining}. \cref{fig:implict_corruptions} visualizes the corruptions generated by $\mathcal{M}_{32}$ on a test image of \textsc{cifar10} under different $\lambda$. We can see that $\lambda$ controls the severity of the corruptions, with smaller $\lambda$ corresponding to higher severity.

\paragraph{Is a model robust against its own corruptions?} We use both models $\mathcal{M}_{16}$ and $\mathcal{M}_{32}$ to generate corruptions on a subset of  $1024$ test images of \textsc{cifar10}. We generate $8$ corruptions per test image. The left plot of \cref{fig:cross_testing_implicit_corruption} shows that each model is robust against its own implicit corruptions even when the corruption is severe, as evidenced by the small performance degradation under different $\lambda$.
By comparing the right plot to the left plot, we can see that each model is less resistant to the corruptions generated by the other model than its own corruptions.
Crucially, however, the performance of $\mathcal{M}_{32}$ under the corruptions generated by $\mathcal{M}_{16}$ is better than the reverse. 
This example thus suggests that while each model is resistant to its own corruptions, the model with higher entropy shows better robustness against the corruptions created by the other model.\footnote{We note that this a proof of concept and more experiments are needed to verify if these results hold true in general.}
\section{Maximizing variational entropy}\label{sec:maximize_entropy}

The previous sections motivated the usage of variational posteriors with high entropy from the perspective of simulating a diverse set of input corruptions. In this section, we discuss a simple method to increase the variational entropy.

\subsection{The augmented ELBO}
\label{sec:augmented_elbo}

Our goal is to find posterior approximations that have high entropy. In the previous section, we considered a heuristic approach of initializing $q(\Z)$ with high entropy  (\cref{fig:entropy_plot}). 
However, if the initial entropy of $q(\Z)$ is too high, training will converge slowly due to high variance in the gradients.

Here we consider the approach of augmenting the original ELBO in \cref{eq:elbo} with an extra $\gamma$-weighted entropy term, adapting \citet{mandt2016variational}.
The augmented $\gamma$-ELBO is
\begin{align}
    &\L_\gamma(\htheta, \phi) = \L(\htheta, \phi) + \gamma \H\big[q_\phi(\Z)\big]\\
	&= \underbrace{\E_{q_{\phi}(\Z)}\Big[\log p(\D | \htheta,\Z)\Big]}_{\text{expected log-likelihood}}
	- \underbrace{\H\big[q_{\phi}(\Z), p(\Z)\big]}_{\text{cross-entropy}} \\
	& \qquad \qquad \quad+ \underbrace{(\gamma+1) \H\big[q_\phi(\Z)\big]}_{\text{variational entropy}} + \underbrace{\log p(\htheta)}_{\text{weight prior}}, \label{eq:entropy_elbo}
\end{align}
where we decompose the KL into its cross-entropy and entropy terms. $\gamma \ge 0$ controls the amount of extra entropy, with $\gamma=0$ reducing to the classic ELBO in \cref{eq:elbo}. We can interpret the terms in \cref{eq:entropy_elbo} as follows: the first term fits the variational parameters to the dataset; the second and fourth terms regularize $\phi$ and $\htheta$ respectively; the third term increases the entropy of the variational posterior.

\subsection{Tempered posterior inference}
\begin{figure}[t]
    \centering
    \begin{subfigure}[c][3.6cm][c]{.40\columnwidth}
        \includegraphics[height=3.0cm]{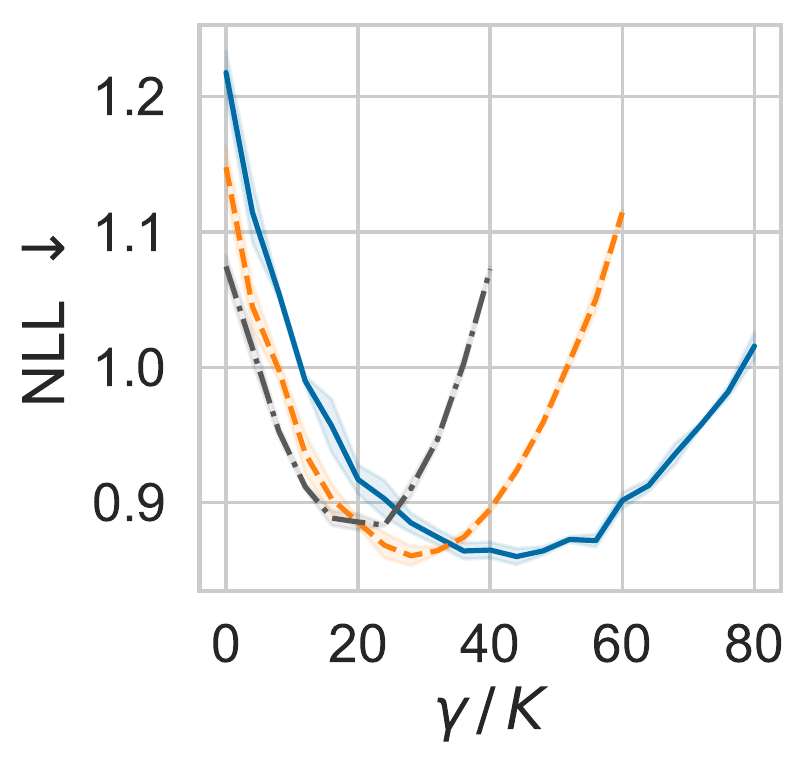}
        \caption{Validation}
        \label{fig:vgg16_cifar100_K_valid}
    \end{subfigure}
    \begin{subfigure}[c][3.6cm][c]{.40\columnwidth}
        \includegraphics[height=3.0cm]{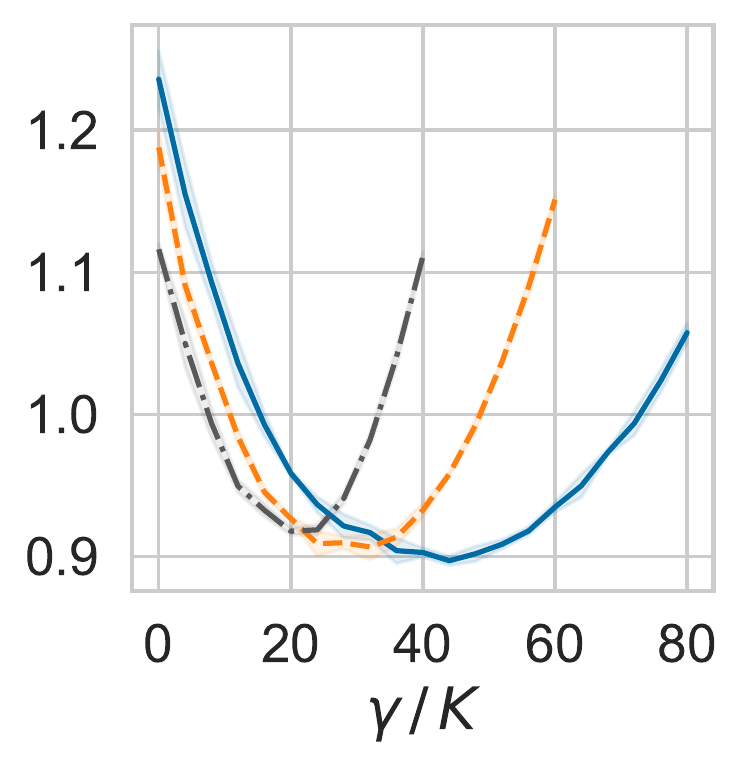}
        \caption{Test}
        \label{fig:vgg16_cifar100_K_test}
    \end{subfigure}
    \begin{subfigure}[c][3.6cm][c]{.40\columnwidth}
        \includegraphics[height=3.0cm]{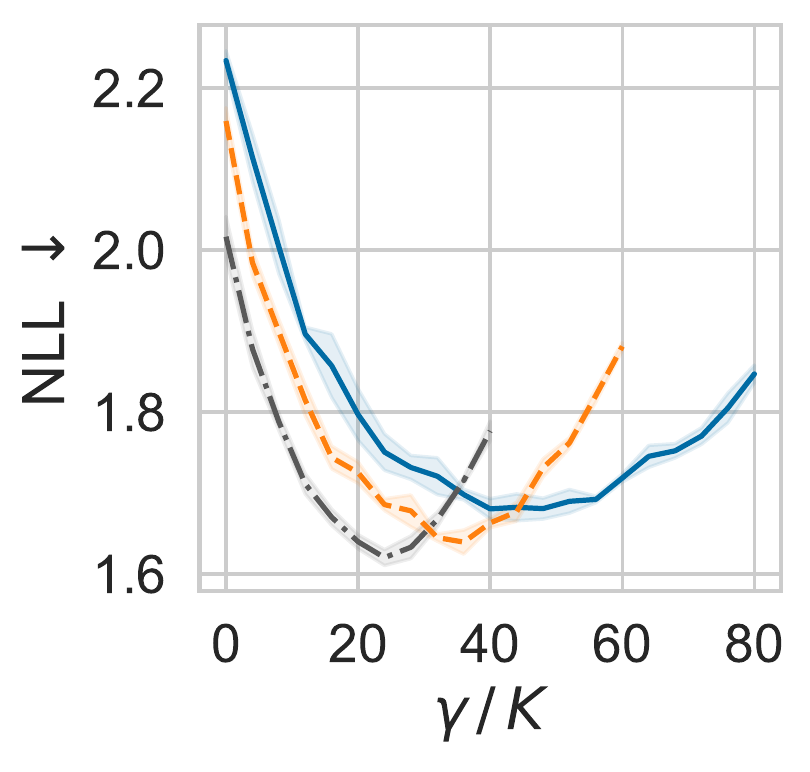}
        \caption{Corruption 1, 2, 3}
        \label{fig:vgg16_cifar100_K_123}
    \end{subfigure}
    \begin{subfigure}[c][3.6cm][c]{.40\columnwidth}
        \includegraphics[height=3.0cm]{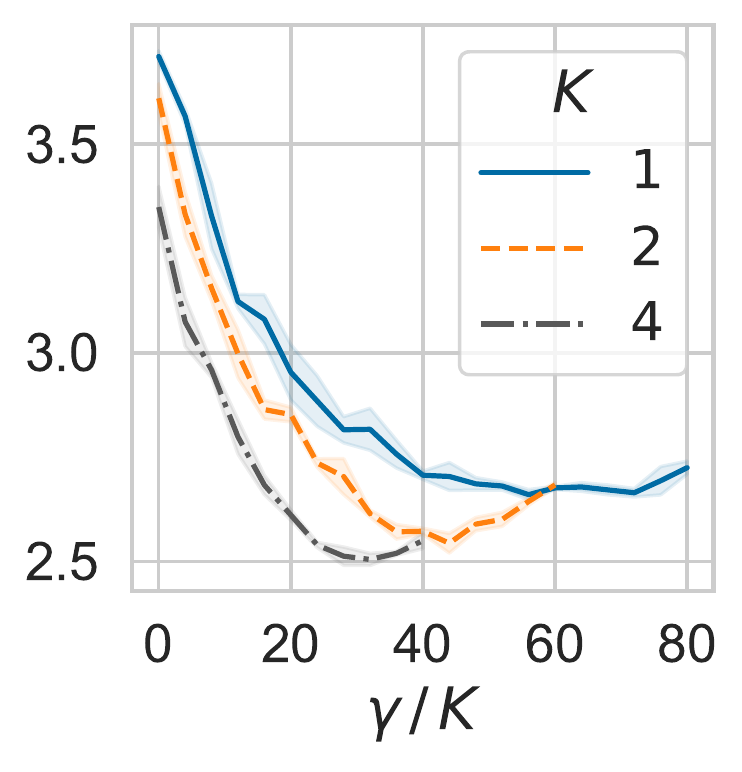}
        \caption{Corruption 4, 5}
        \label{fig:vgg16_cifar100_K_45}
    \end{subfigure}
    \caption{Results of (\textsc{vgg16} / \textsc{cifar100} / out) with different $K$. The results in (c) are averaged over the first three levels of corruption, and those in (d) are averaged over the last two levels.
    Notice that we rescale $\gamma$ by $K$ in the x-axis to provide better visualization, as we find that larger $K$ requires higher optimal $\gamma$. 
    We report the mean and standard deviation over 5 runs for each result.
    Overall, more components provide better optimal performance on OOD data. Higher $\gamma$ provides better OOD performance as the cost of ID performance.}
    \label{fig:vgg16_cifar100_K}
\end{figure}
\begin{figure}[t]
    \centering
    \begin{subfigure}[c][3.6cm][c]{.40\columnwidth}
        \includegraphics[height=3.0cm]{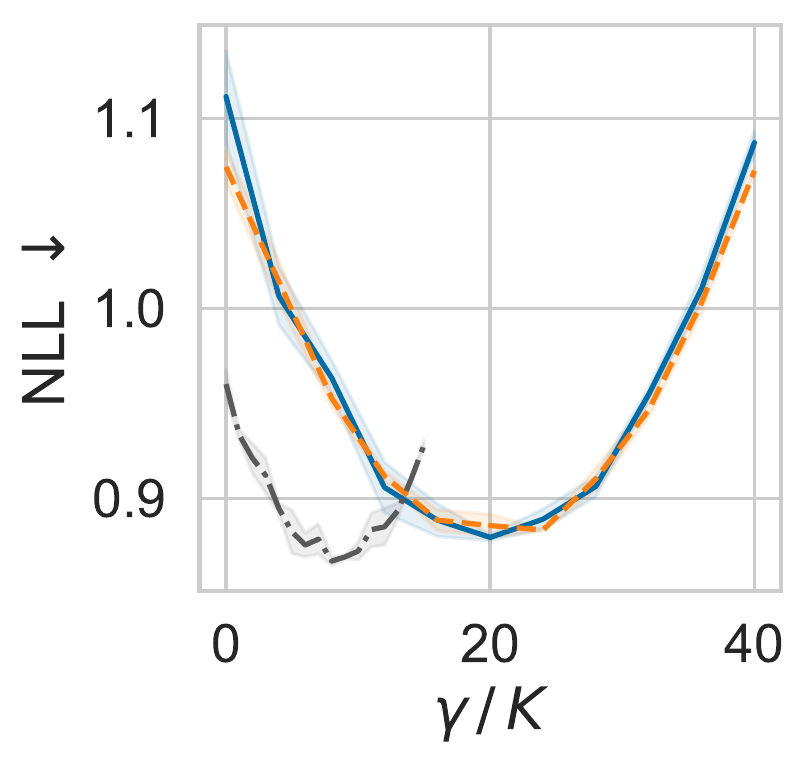}
        \caption{Validation}
    \end{subfigure}
    \begin{subfigure}[c][3.6cm][c]{.40\columnwidth}
        \includegraphics[height=3.0cm]{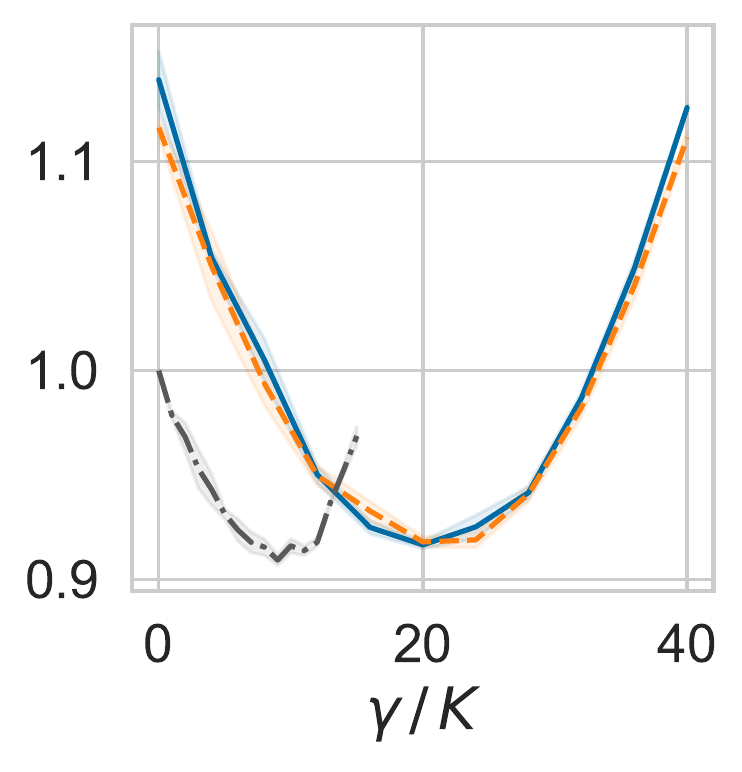}
        \caption{Test}
    \end{subfigure}
    \begin{subfigure}[c][3.6cm][c]{.40\columnwidth}
        \includegraphics[height=3.0cm]{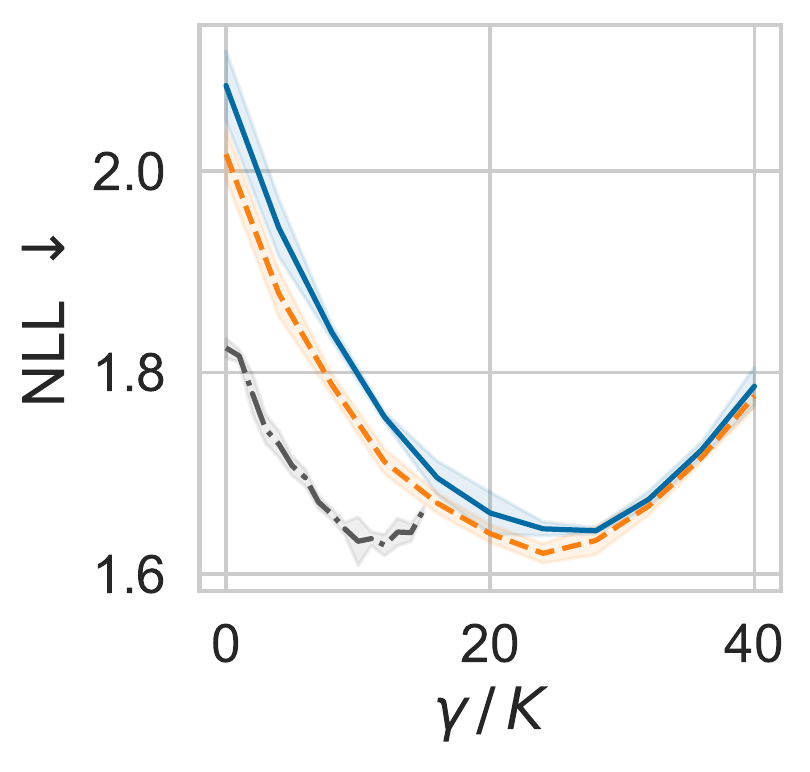}
        \caption{Corruption 1, 2, 3}
    \end{subfigure}
    \begin{subfigure}[c][3.6cm][c]{.40\columnwidth}
        \includegraphics[height=3.0cm]{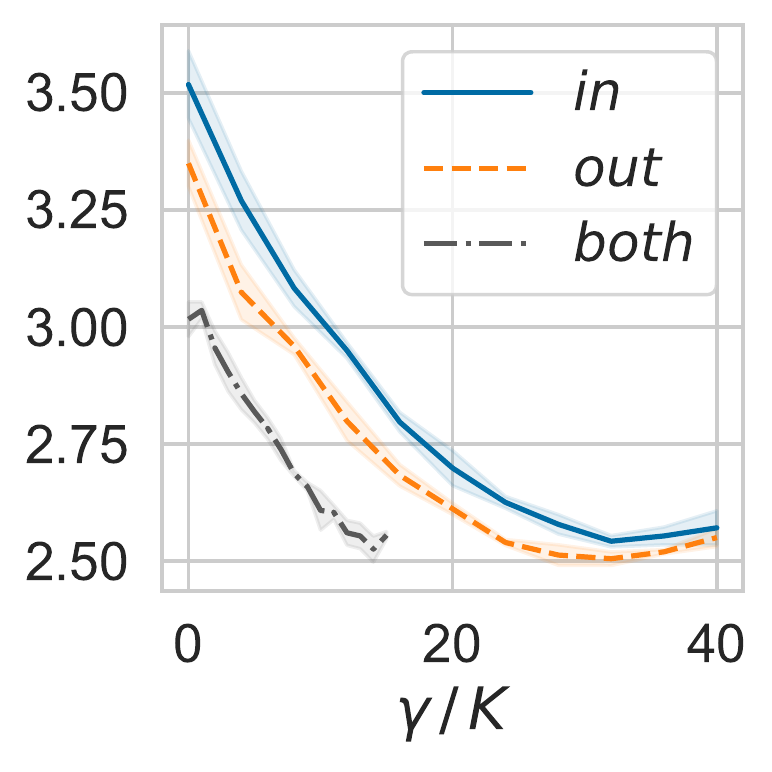}
        \caption{Corruption 4, 5}
    \end{subfigure}
    \caption{Results of \textsc{vgg16} on \textsc{cifar100} with different latent variable structures. Here we use $K=4$ components. We report the mean and standard deviation over 5 runs for each result. Overall, using either only the latent input variables or latent output variables requires higher optimal $\gamma$ than using both. Using only the latent output variables produces better results than the latent input variables on OOD data, despite similar ID performance.}
    \label{fig:vgg16_cifar100_noise}
\end{figure}

One could also arrive at \cref{eq:entropy_elbo} by minimizing the KL divergence between the approximate posterior $q_{\phi}(\htheta,\Z)$ and the tempered posterior $p_\gamma(\theta, \Z|\D)$ \citep{mandt2016variational}:
\begin{align}\label{eq:tempered_posterior}
	p_\gamma(\theta, \Z|\D) &= \frac{p(\D|\theta, \Z)^{\tau}p(\Z,\theta)^{\tau}}{p_\gamma(\D)} \\
    p_\gamma(\D) &= \int_{\theta}\int_{\Z} p(\D|\theta, \Z)^{\tau}p(\Z,\theta)^{\tau} d\Z d\theta,
\end{align}
where the temperature $\tau = 1/(\gamma+1)$.
The tempered posterior variational approximation
\begin{align}
	\argmin_{\htheta,\phi} \:\: \frac{1}{\tau} \, \kl\left[ q_{\phi}(\hat{\theta}, \Z) \, \left|\right| \, p_\gamma(\theta, \Z|\D) \right] 
\end{align}
is equivalent to tempered ELBO maximization
\begin{align}
    \argmax_{\htheta, \phi} \: \L_\gamma(\htheta, \phi) - \log p_\gamma(\D)^{\frac{1}{\tau}}.
\end{align}
We refer the reader to \cref{appendix:tempered_elbo} for detailed derivations. 
The entropy-regularized $\gamma$-ELBO thus corresponds to the family of tempered variational inference, and with positive $\gamma > 0$, to `hot' posteriors \citep{wenzel2020good}. In the next section, we will demonstrate empirically the benefits of such hot posteriors in node-based BNNs.













\section{Experiments}
\label{sec:experiments}

In this section, we present experimental results of node-based BNNs on image classification tasks. For the datasets, we use \textsc{cifar} \cite{krizhevsky2009cifar} and \textsc{tinyimagenet} \cite{Le2015TinyIV}, which have corrupted versions of the test set provided by \citet{hendrycks2019robustness}. We use \textsc{vgg16} \cite{simonyan2014very}, \textsc{resnet18} \cite{he2016deep} and \textsc{preactresnet18} \cite{he2016identity} for the architectures. We test three structures of latent variables: \emph{in}, where we only use the input latent variables $\{\z^\ell\}_{\ell=1}^L$; \emph{out}, where we only use the output latent variables $\{\s^\ell\}_{\ell=1}^L$; and \emph{both}, where we use both $\{\z^\ell\}_{\ell=1}^L$ and $\{\s^\ell\}_{\ell=1}^L$. We use $K \in \{1, 2, 4\}$ Gaussian component(s) in the variational posterior. For each result, we report the mean and standard deviation over multiple runs.

\subsection{Effects of $\gamma$ on covariate shift}

In this section, we study the changes in performance of the model trained with the $\gamma$-ELBO objective as we increase $\gamma$.
We perform experiments with \textsc{vgg16} on \textsc{cifar100}, and use the corrupted test set of \textsc{cifar100} provided by \citet{hendrycks2019robustness}.
In \cref{fig:vgg16_cifar100_K}, we show the \emph{out} model's behaviour under a different number of Gaussian components $K$.
In \cref{fig:vgg16_cifar100_noise}, we show the results of a model with $K=4$ components under the different latent variable structures \emph{in}, \emph{out}, and \emph{both}.

These figures show that performance across different test sets improves as $\gamma$ increases up until a threshold and then degrades afterward.
The optimal $\gamma$ for each set of test images correlates with the severity of the corruptions, where more severe corruptions can be handled by enforcing more diverse set of implicit corruptions during training.
However, learning on a more diverse implicit corruptions requires higher capacity, and reduces the learning capacity needed to obtain good performance on the in-distribution (ID) data. 
The entropy coefficient $\gamma$ thus controls the induced trade-off between ID performance and out-of-distribution (OOD) robustness.

\cref{fig:vgg16_cifar100_K} shows that for ID data, the optimal performance of the model (at optimal $\gamma$) remains similar under different $K$.
On OOD data, however, higher $K$ consistently produces better results as $\gamma$ varies.
The optimal $\gamma$ is higher for variational distributions with more components.
This finding is likely because with more mixture components, the variational posterior can approximate the true posterior more accurately, and thus it can better expand into the high-density region of the true posterior as its entropy increases.

\cref{fig:vgg16_cifar100_noise} shows the optimal performance on ID data is quite similar between different latent architectures.
On OOD, the optimal performance of using both input and output latent variables is similar to using only output latent variables, while using only input latent variables produces slightly worse optimal performance.
The optimal $\gamma$ is lower when the model uses both types of latent variables ($\z,\s$), because the entropy of the product of two latent variables increases rapidly as we increase the entropy of both latent variables.

We also observe these patterns in other architectures and datasets (see \cref{app:gamma_effect}).
In summary, from our experimental results we find that using only output latent variables with a sufficient number of components (e.g., $K=4$) achieves excellent results for node-based BNNs in our benchmark.

\subsection{Effects of $\gamma$ on robustness against noisy labels}
\begin{figure}[t]
    \centering
    \begin{subfigure}[c][3.5cm][c]{.32\columnwidth}
        \includegraphics[height=3.0cm]{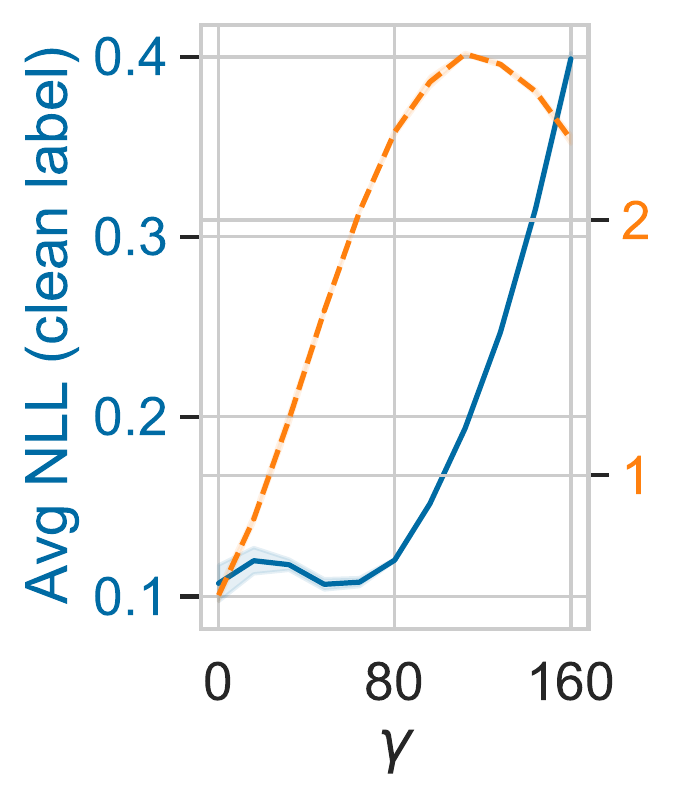}
        \caption{$20\%$}
        \label{fig:nll_clean_noisy_label20}
    \end{subfigure}
    \hfill
    \begin{subfigure}[c][3.5cm][c]{.30\columnwidth}
        \includegraphics[height=3.0cm]{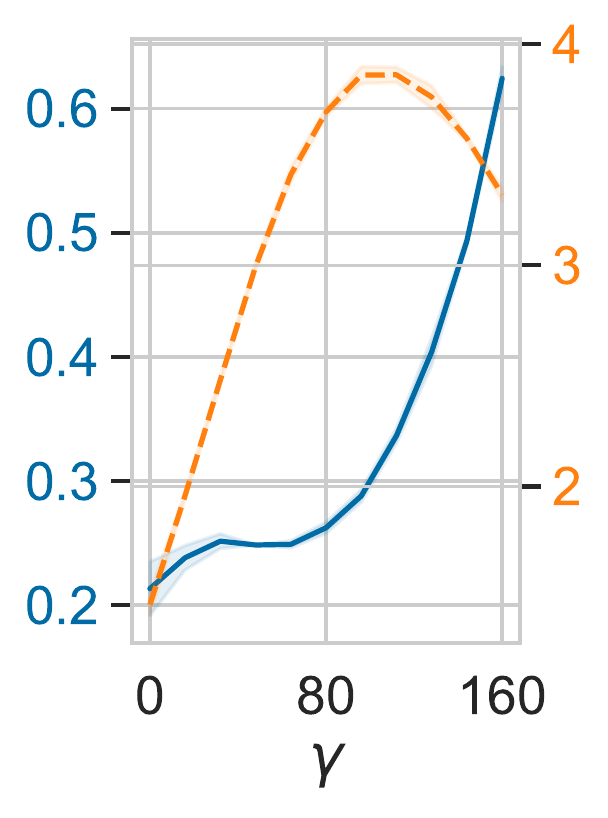}
        \caption{$40\%$}
        \label{fig:nll_clean_noisy_label40}
    \end{subfigure}
    \hfill
    \begin{subfigure}[c][3.5cm][c]{.32\columnwidth}
        \includegraphics[height=3.0cm]{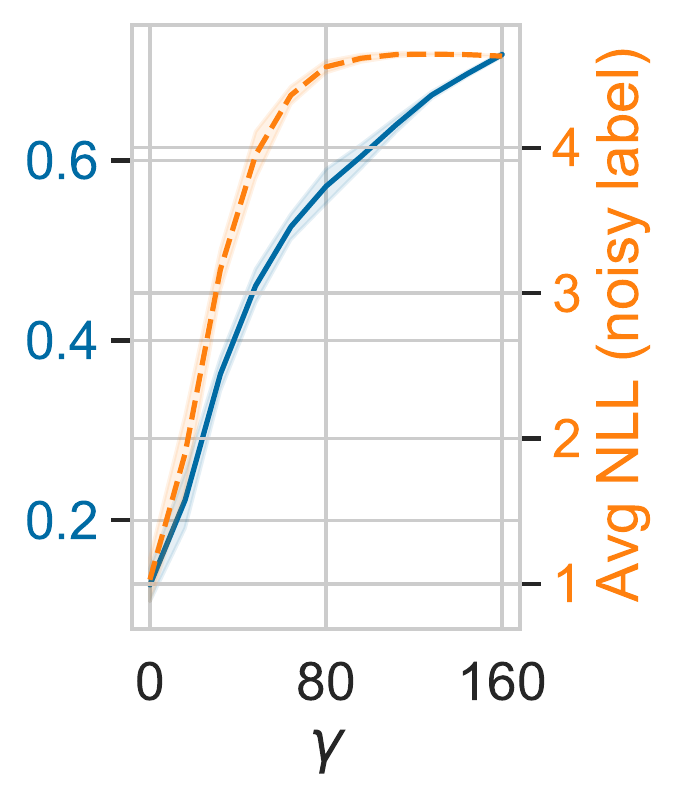}
        \caption{$80\%$}
        \label{fig:nll_clean_noisy_label80}
    \end{subfigure}
    \caption{Results of \textsc{resnet18} on two subsets of \textsc{cifar10} training samples with clean and noisy labels. Here we use $K=4$ components and only the latent output variables. We denote the percentage of training samples with corrupted labels under each subfigure. We report the mean and standard deviation over 5 runs for each result. As $\gamma$ increases, the NLL of noisy labels increases much faster than that of clean labels even when the majority of labels are wrong (c), indicating that higher $\gamma$ prevents the model from memorizing random labels.}
    \label{fig:nll_clean_noisy_label}
\end{figure}
\begin{figure}[t]
    \centering
    \begin{subfigure}[c][4.0cm][c]{.35\columnwidth}
        \includegraphics[height=3.5cm]{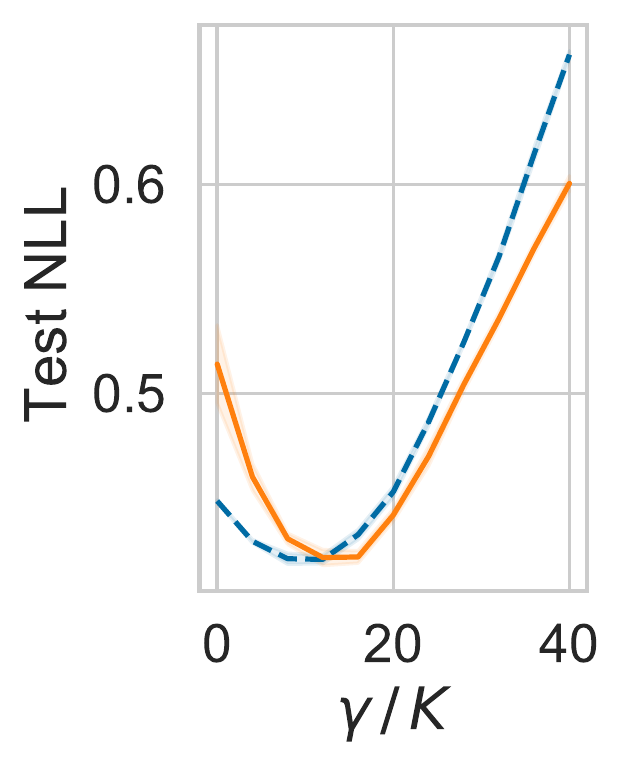}
        \caption{$20\%$}
    \end{subfigure}
    \begin{subfigure}[c][4.0cm][c]{.31\columnwidth}
        \includegraphics[height=3.5cm]{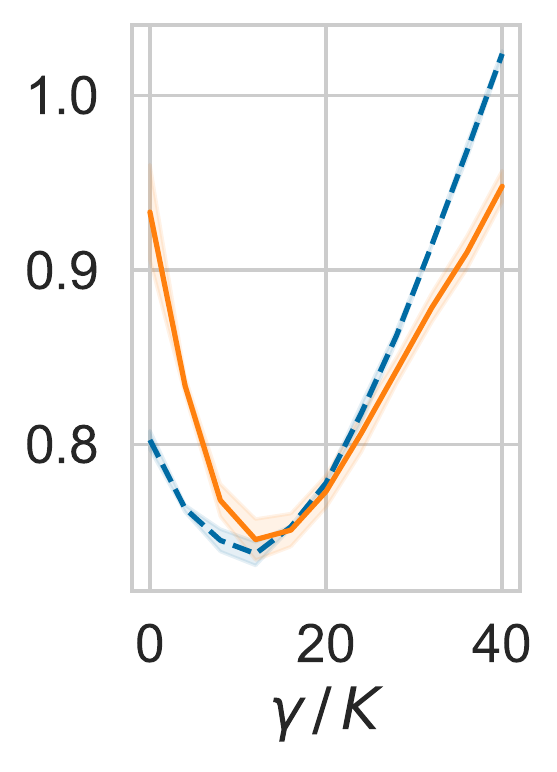}
        \caption{$40\%$}
    \end{subfigure}
    \begin{subfigure}[c][4.0cm][c]{.31\columnwidth}
        \includegraphics[height=3.5cm]{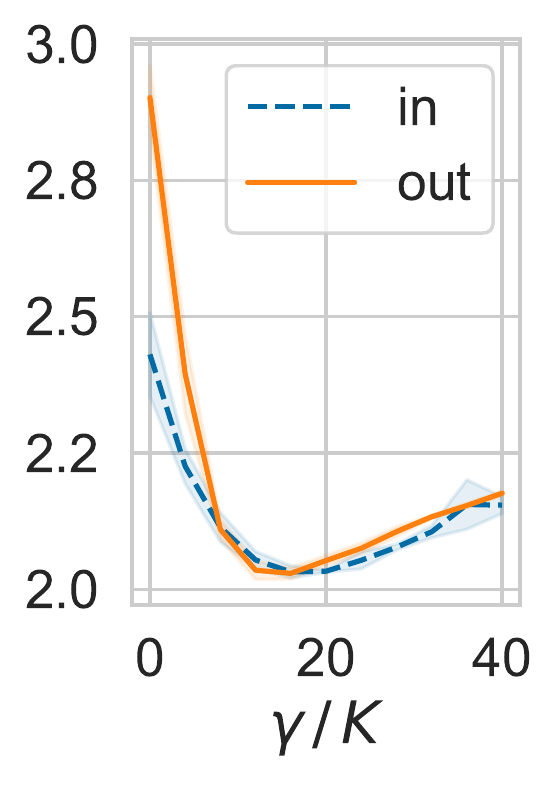}
        \caption{$80\%$}
    \end{subfigure}
    \caption{Results of \textsc{resnet18} on clean \textsc{cifar10} test sets under different percentages of noise in training labels. We report the mean and standard deviation over 5 runs for each result. As high $\gamma$ prevents learning from noisy labels as demonstrated in \cref{fig:nll_clean_noisy_label}, it leads to improved performance on clean test sets.}
    \label{fig:resnet18_cifar10_noise}
\end{figure}

Learning wrong labels amounts to memorizing random patterns, which requires more capacity from the model than learning generalizable patterns \cite{arpit17a}.
We hypothesize that if we corrupt wrongly labelled training samples with sufficiently diverse implicit corruptions, we overwhelm the neural network making it unable to memorize these spurious patterns during training.
To test this intuition, we follow the experiment in \citet{jiang2018mentornet}, where we take a percentage of training samples in \textsc{cifar10} and corrupt their labels.
We thus split the training set into two parts: $\D_1$ containing only samples with correct labels, and $\D_2$ including those with wrong labels.
We then track the final NLL of $\D_1$ and $\D_2$ under different $\gamma$, and visualize the results in \cref{fig:nll_clean_noisy_label}.
This figure shows that as $\gamma$ increases, the NLL of $\D_2$ (noisy labels) increases much faster than that of $\D_1$ (clean labels), indicating that the network fails to learn random patterns under simulated corruptions.
As a consequence, the model generalizes better on the test set, as shown in \cref{fig:resnet18_cifar10_noise}.

\subsection{Benchmark results}
\begin{figure}[t]
    \centering
    \includegraphics[width=\columnwidth]{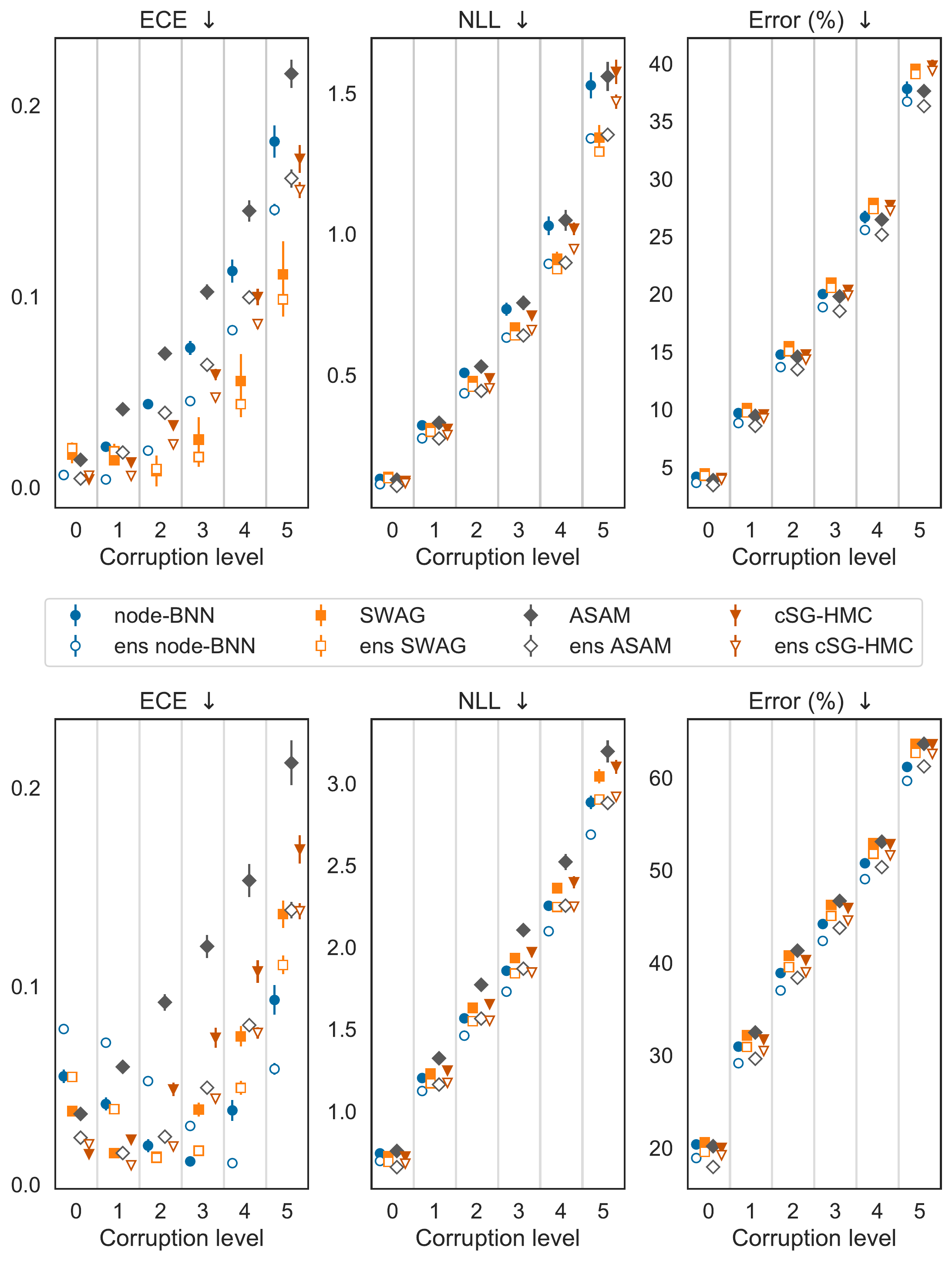}
    \caption{Results of \textsc{resnet18} on \textsc{cifar10} \textbf{(top)} and \textsc{cifar100} \textbf{(bottom)}. We use $K=4$ and only the latent output variables for node-based BNNs. We plot ECE, NLL and error for different corruption levels, where level $0$ indicates no corruption. We report the average performance over 19 corruption types for level 1 to 5. We denote the ensemble of a method using the shorthand \emph{ens} in front of the name. Each result is the average over 25 runs for \emph{non-ens} versions and 5 runs for \emph{ens} versions. The error bars represent the standard deviations across different runs. Node-based BNNs and their ensembles (\textcolor[RGB]{0,107,164}{blue}) perform best across all metrics on OOD data of \textsc{cifar100}, while having competitive results on \textsc{cifar10}. We include a larger version of this plot in \cref{app:benchmark}.}
    \label{fig:benchmark_resnet18}
\end{figure}
\begin{figure}[t]
    \centering
    \includegraphics[width=\columnwidth]{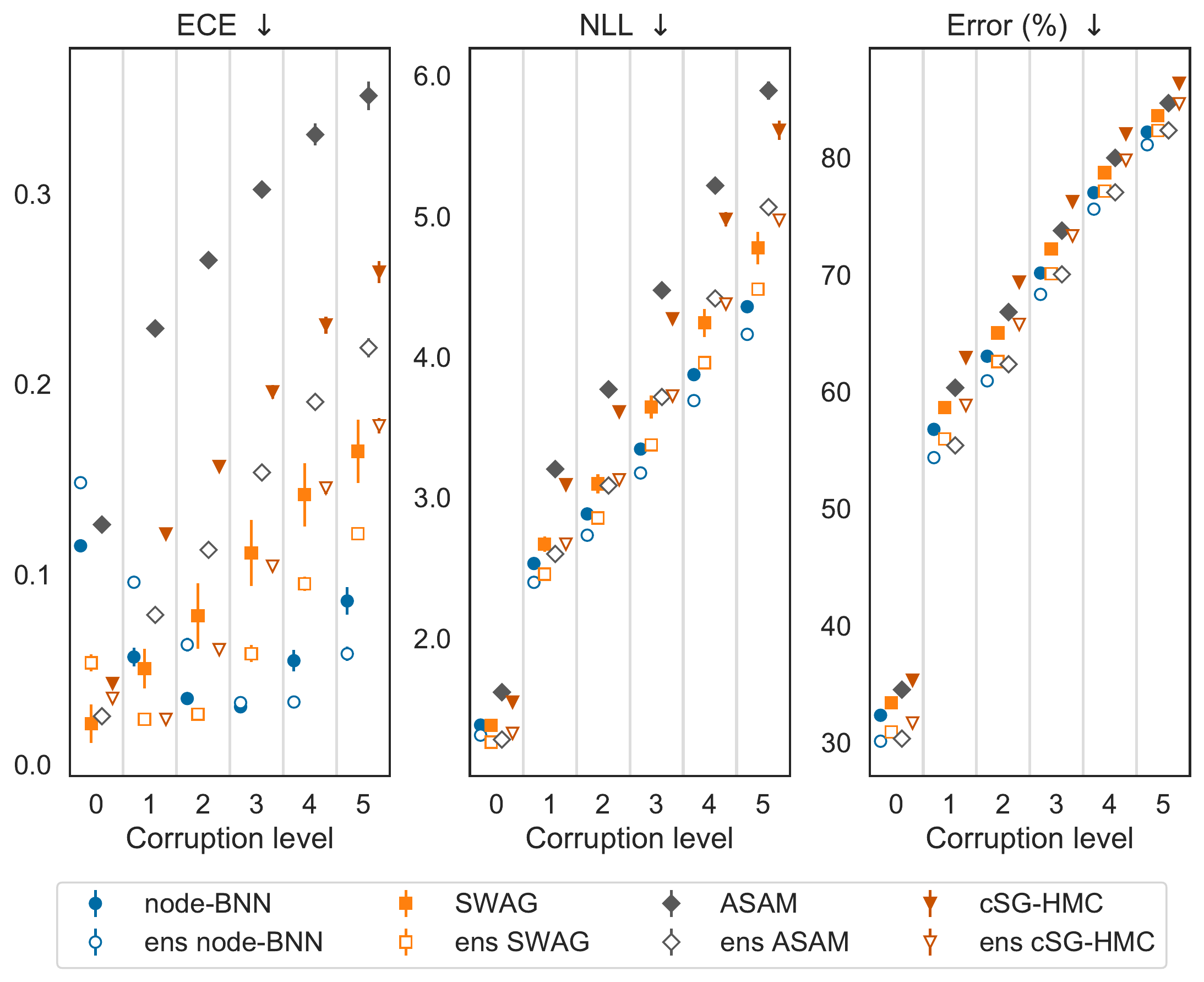}
    \caption{Results of \textsc{preactresnet18} on \textsc{tinyimagenet}. We use $K=4$ and only the latent output variables for node-based BNNs. We plot ECE, NLL and error for different corruption levels, where level $0$ indicates no corruption. We report the average performance over 19 corruption types for level 1 to 5. We denote the ensemble of a method using the shorthand \emph{ens} in front of the name. Each result is the average over 25 runs for \emph{non-ens} versions and 5 runs for \emph{ens} versions. The error bars represent the standard deviations across different runs. Node-based BNNs and their ensembles (\textcolor[RGB]{0,107,164}{blue}) perform best accross all metrics on OOD data, while having competitive performance on ID data. We include a larger version of this plot in \cref{app:benchmark}.}
    \label{fig:benchmark_preeacresnet18}
\end{figure}

\cref{fig:benchmark_resnet18,fig:benchmark_preeacresnet18} present the results of node-based BNNs and baselines on \textsc{cifar10}/\textsc{cifar100} and \textsc{tinyimagenet}.
We choose SWAG \cite{maddox_2019_simple}, cSG-HMC \cite{zhang2020csgmcmc} and ASAM \cite{kwon2021asam} as our baselines.
These are strong baselines, as both SWAG and cSG-HMC have demonstrated state-of-the-art uncertainty estimation, while ASAM produce better MAP models than stochastic gradient descent by actively seeking wide loss valleys.
We repeated each experiment 25 times with different random seeds.
For each method, we also consider its ensemble version where we combine 5 models from different runs when making predictions.
For the ensemble versions, each experiment is repeated 5 times.
We use 30 Monte Carlo samples for node-based BNNs, SWAG, cSG-HMC and their ensemble versions to estimate the posterior predictive distribution.
We use standard performance metrics of expected calibration error (ECE) \cite{Naeini2015}, NLL and predictive error.
We use \textsc{resnet18} for \textsc{cifar10}/\textsc{cifar100} and \textsc{preactresnet18} for \textsc{tinyimagenet}.
We also include the result of \textsc{vgg16} on \textsc{cifar10}/\textsc{cifar100} in \cref{app:benchmark}.
For evaluation, we use the corrupted test images provided by \citet{hendrycks2019robustness}.

On \textsc{cifar100}, node-based BNNs outperform the baselines in NLL and error, however SWAG performs best on \textsc{cifar10}.
Interestingly, in \textsc{cifar100}, node-based BNNs and their ensembles have worse ECE than the baselines on ID data, however as the test images become increasingly corrupted, the ECEs of the baselines degrade rapidly while the ECE of node-based BNNs remains below a threshold.
Similar behaviors are observed on \textsc{tinyimagenet}, with the node-based BNNs produce the lowest NLL and error while not experiencing ECE degradation under corruptions.



\section{Related works}

\paragraph{Multiplicative latent node variables in BNNs.} There have been several earlier works that utilize multiplicative latent node variables, either as a primary source of predictive uncertainty such as MC-Dropout \cite{gal2016dropout}, Variational Dropout \cite{variational_dropout}, Rank-1 BNNs \cite{dusenberry20a} and Structured Dropout \cite{nguyen2021structured}; or to improve the flexibility of the mean-field Gaussian posterior in variational inference \cite{louizos2017multiplicative}. Here we study the contribution of these latent variables to robustness under covariate shift.

\paragraph{BNNs under covariate shift.} Previous works have evaluated the predictive uncertainty of BNNs under covariate shift \cite{ovadia2019can, izmailov2021bayesian}, with the recent work by \citet{izmailov2021bayesian} showing that standard BNNs with high-fidelity posteriors perform worse than MAP solutions under covariate shift. \citet{izmailov2021dangers} attributed this phenomenon to the absence of posterior contraction on the null-space of the data manifold. This problem is avoided in node-based BNNs as they still maintain a point-estimate for the weights.

\paragraph{Dropout as data augmentation.} Similar to our study, a previous work by \citet{bouthillier2015dropout} studied Dropout from the data augmentation perspective. Here we study latent variables with more flexible posterior (mixture of Gaussians) and focus on simulating input corruptions for OOD robustness.

\paragraph{Adversarial robustness via feature perturbations.} Data-space perturbations have been investigated as a means to defend neural networks against adversarial attacks \cite{li2018certified, jeddi2020learn2perturb, vadera2020assessing}.

\paragraph{Tempered posteriors.} Tempered posteriors have been used in variational inference to obtain better variational posterior approximations \cite{mandt2016variational}. A recent study put the focus on the cold posterior effect of weight-based BNNs \cite{wenzel2020good}. We have shown that our approach of regularizing the variational entropy is equivalent to performing variational inference with a hot posterior as the target distribution.
Tempered posteriors have also been studied in Bayesian statistics as a means to defend against model misspecification \cite{grunwald2012safe,miller_robust_2019,alquier2020concentration,medina2021robustness}. Covariate shift is a form of model misspecification, as model mismatch arises from using a model trained under different assumptions about the statistics of the data.

\section{Conclusion}
We analyzed node-based BNNs from the perspective of using latent node variables for simulating input corruptions.
We showed that by regularizing the entropy of the latent variables, we increase the diversity of the implicit corruptions, and thus improve performance of node-based BNNs under covariate shift.
Across \textsc{cifar10}, \textsc{cifar100} and \textsc{tinyimagenet}, entropy regularized node-based BNNs produce excellent results in uncertainty metrics on OOD data.

In this study, we focused on variational inference, leaving the study of implicit corruptions under other approximate inference methods as future work. 
Furthermore, our work shows the benefits of hot posteriors and argues for an inherent trade-off between ID and OOD performance in node-based BNNs. 
It is an interesting future direction to study these questions in weight-based BNNs.
Finally, our work presented entropy as a surprisingly useful summary statistic that can partially explain the complex connection between the variational posterior and corruption robustness. 
One important research direction is to develop more informative statistics that can better encapsulate this connection.


\section*{Acknowledgement}
This work was supported by the Academy of Finland (Flagship programme: Finnish Center for Artificial Intelligence FCAI and grants no. 292334, 294238, 319264, 328400) and UKRI Turing AI World-Leading Researcher Fellowship, EP/W002973/1. We acknowledge the computational resources provided by Aalto Science-IT project and CSC–IT Center for Science, Finland.

\bibliography{references}
\bibliographystyle{icml2022}

\newpage
\appendix
\onecolumn
\section{Original ELBO derivation}\label{appendix:orig_elbo}
Here we provide a detail derivation of the ELBO in \cref{eq:elbo}.
We assume a prior $p(\theta, \Z)=p(\theta)p(\Z)$ for the parameters $\theta$ and latent variables $\Z$, and we assume a variational posterior $q_{\phi, \htheta}(\theta, \Z)=\delta(\theta-\htheta)q_\phi(\Z)$ where $\delta(.)$ is a Dirac delta distribution. We arrive at the ELBO in \cref{eq:elbo} by minimizing the KL divergence between the variational approximation and the true posterior with respect to the variational parameters $(\htheta, \phi)$:
\begin{align}
	&\argmin_{\phi, \htheta} \kl\left[q_{\phi, \htheta}(\theta, \Z)||p(\theta, \Z|\D)\right] \\
	&= \argmin_{\phi, \htheta} \E_{q_{\phi, \htheta}(\theta, \Z)}\left[\log q_{\phi, \htheta}(\theta, \Z) - \log p(\D|\theta,\Z) - \log p(\theta, \Z) + \log p(\D)\right] \\
	&= \argmin_{\phi, \htheta} \E_{q_{\phi}(\Z)}\left[-\log p(\D|\htheta,\Z)\right]
	+ \kl\left[q_{\phi}(\Z)||p(\Z)\right] - \log p(\htheta) + \log p(\D) \\
	&= \argmin_{\phi, \htheta} -\L(\htheta, \phi)
\end{align}

\section{Tempered ELBO derivation}\label{appendix:tempered_elbo}
Here we show a connection between the tempered posterior with temperature $\tau=1/(\gamma+1)$ in \cref{eq:tempered_posterior} and the augmented ELBO in Section \ref{sec:augmented_elbo}:
\begin{align}
	&\argmin_{\phi, \htheta} \frac{1}{\tau} \kl\left[q_{\phi, \htheta}(\theta, \Z) \, || \, p_\gamma(\theta, \Z|\D)\right] \\
	&= \argmin_{\phi, \htheta} \frac{1}{\tau} \E_{q_{\phi, \htheta}(\theta, \Z)}\left[\log q_{\phi, \htheta}(\theta, \Z) - \tau\log p(\D|\theta,\Z) - \tau\log p(\theta, \Z) + \log p_\gamma(\D)\right] \\
	&= \argmin_{\phi, \htheta} \E_{q_{\phi, \htheta}(z, \theta)}\left[\frac{1}{\tau}\log q_{\phi, \htheta}(\theta, \Z) - \log p(\D|\theta,\Z) - \log p(\theta) - \log p(\Z) \right] + \frac{1}{\tau}\log p_\gamma(\D) \\
	&= \argmin_{\phi, \htheta} -\E_{q_{\phi}(\Z)}\left[\log p(\D|\htheta,\Z)\right]
	+ \kl\left[q_{\phi}(\Z)||p(\Z)\right] - \gamma \H\left[q_\phi(\Z)\right] - \log p(\htheta) + \frac{1}{\tau}\log p_\gamma(\D) \\
	&= \argmin_{\phi, \htheta} -\L_\gamma(\htheta, \phi) + \log p_\gamma(\D)^{\frac{1}{\tau}}
\end{align}

\section{Derivation of layer-wise activation shifts due to input corruptions}\label{appendix:activation_shifts}
Here we explain in detail the approximation of layer-wise activation shifts in \cref{eq:shift_ell_approx}.
To simulate covariate shift, one can take an input $\x$ assumed to come from the same distribution as the training samples and apply a corruption $\g^0$ to form a shifted version $\x^c$ of $\x$:
\begin{equation}
    \x^c \triangleq \x + \g^0(\x)
\end{equation}
For instance, $\x$ could be an image and $\g^0$ can represent the shot noise corruption as seen in \citet{hendrycks2019robustness}.
The corruption $\g^0(\x)$ creates a shift in the activation of the first layer $\f^1$ which can be approximated using the first-order Taylor expansion:
\begin{align}
    \g^1(\x) &= \f^1(\x^c) - \f^1(\x) \\
    &= \sigma\left(\W^1(\x+\g^0(\x)) + \b^1\right) - \sigma\left(\W^1\x+ \b^1\right) \\
    &\approx \J_\sigma\left[ \h^1(\x) \right]\left(\W^1\g^0(\x)\right)
\end{align}
where $\J_\sigma=\partial \sigma/\partial \h$ denotes the Jacobian of the activation $\sigma$ with respect to pre-activation outputs $\h$. Similarly, the approximation of the activation shift in the second layer is:
\begin{align}
    \g^2(\x) &= \f^2(\x^c) - \f^2(\x) \\
    &= \sigma\left(\W^2\f^1(\x^c) + \b^2\right) - \sigma\left(\W^2\f^1(\x)+ \b^2\right) \\
    &= \sigma\left(\W^2(\f^1(\x)+\g^1(\x)) + \b^2\right) - \sigma\left(\W^2\f^1(\x)+ \b^2\right) \\
    &\approx \J_\sigma\left[ \h^2(\x) \right ]\left(\W^2\g^1(\x)\right)
\end{align}
Generally, one can approximate the shift in the output of the $\ell$-th layer caused by $\g(\x)$ as:
\begin{equation}
    \g^\ell(\x) = \f^\ell(\x^c) - \f^\ell(\x) \approx \J_\sigma\left[ \h^\ell (\x) \right] \left(\W^{\ell}\g^{\ell-1}(\x)\right)
\end{equation}

\section{Details on small-scale experiments}
For the small-scale experiments in Section \ref{sec:implicit_corruption}, we use the \textsc{all-cnn-c} architecture from \citet{springenberg2014striving}. We describe this architecture in Table \ref{tab:all-cnn-c}. We train the model for $90$ epochs, and only use the output latent variables and a posterior with $1$ Gaussian component for this experiment
\begin{table}[t]
\centering
\caption{The \textsc{all-cnn-c} architecture}
\label{tab:all-cnn-c}
\vskip 0.15in
\begin{tabular}{c}
\toprule
\textsc{all-cnn-c}        \\
\midrule
Input $32 \times 32$ RGB images                                      \\
\midrule
$3 \times 3$ conv. with 96 output filters, ReLU                      \\
$3 \times 3$ conv. with 96 output filters, ReLU                      \\
\midrule
$3 \times 3$ conv. with 96 output filters and stride $r=2$, ReLU   \\
\midrule
$3 \times 3$ conv. with 192 output filters, ReLU                     \\
$3 \times 3$ conv. with 192 output filters, ReLU                     \\
\midrule
$3 \times 3$ conv. with 192 output filters and stride $r=2$, ReLU  \\
\midrule
$3 \times 3$ conv. with 192 output filters, ReLU                     \\
$1 \times 1$ conv. with 10 output filters, ReLU                      \\
\midrule
Global average pooling                                                                \\
10-way softmax        \\                                      
\bottomrule          
\end{tabular}
\end{table}

\section{Additional visualization of outputs at each layer}
In Section \ref{sec:implicit_corruption}, we provide a PCA visualization of the outputs from the last layer of a node-based \textsc{all-cnn-c} BNN on one sample of \textsc{cifar10}. Here we also provide the same visualizations for the first two and the last two layers of the network. We use the same input image as \cref{fig:conv8_distribution_corruption}.
\begin{figure}[t]
    \centering
    \captionsetup[subfigure]{justification=centering}
    \begin{subfigure}[b]{\textwidth}
         \centering
         \includegraphics[width=\textwidth]{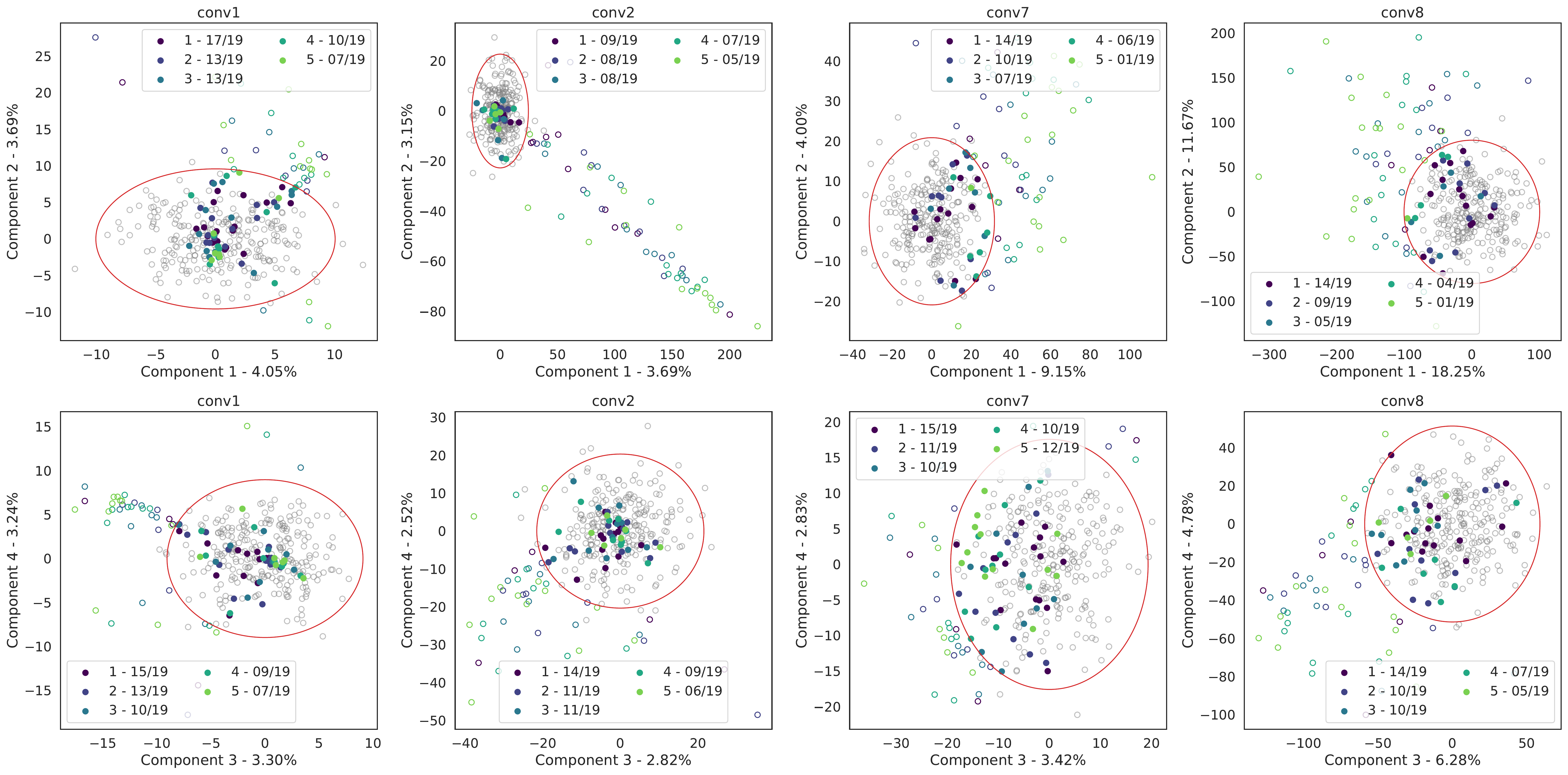}
         \caption{The outputs of the first two and last two layer in $\mathcal{M}_{16}$. $q(\Z)$ is a single Gaussian with the standard deviations initialized from a half normal $\mathcal{N}^+(0.16, 0.02)$}
         \label{fig:activation_pca_std16}
     \end{subfigure}
     \begin{subfigure}[b]{\textwidth}
         \centering
         \includegraphics[width=\textwidth]{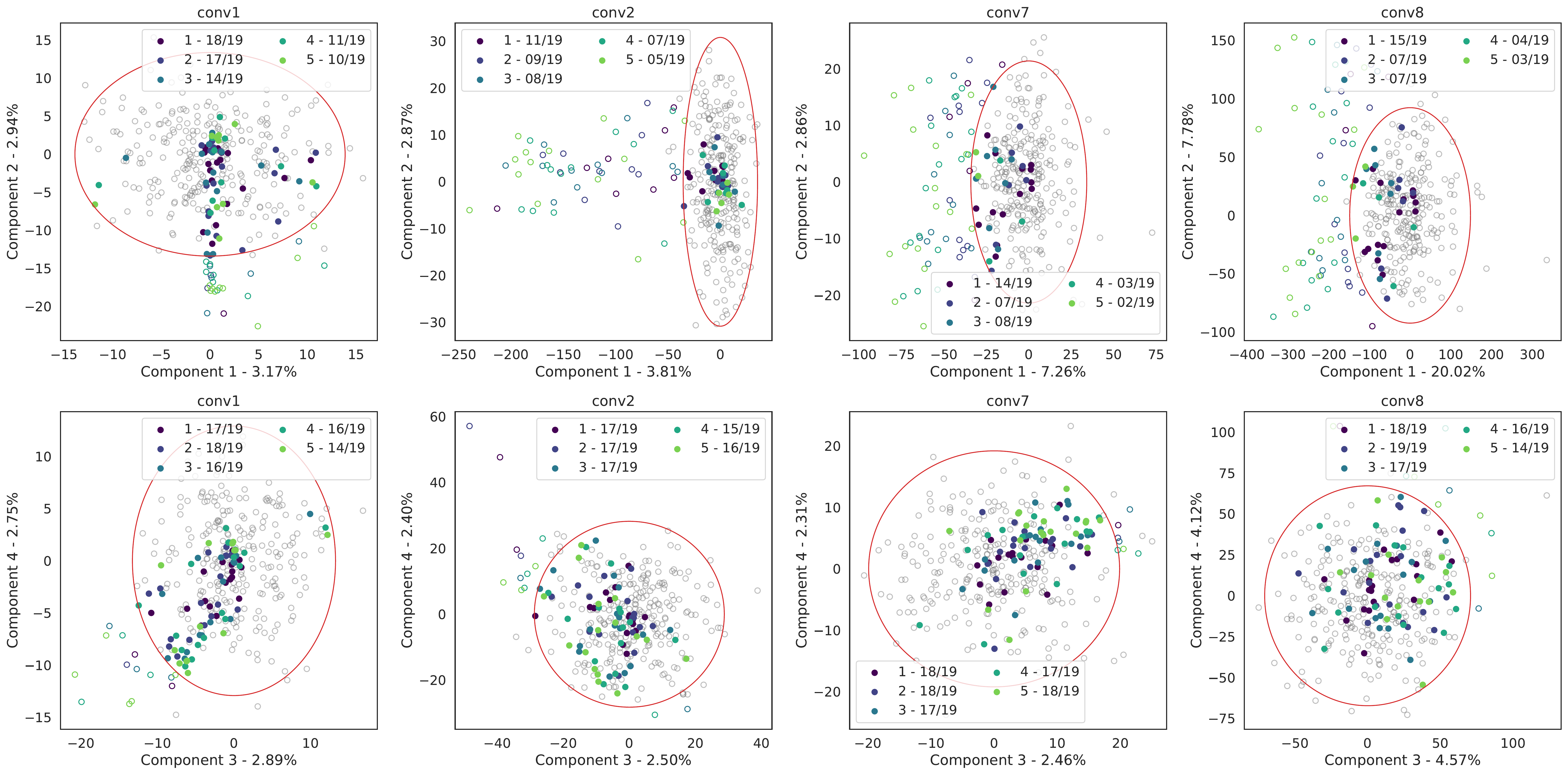}
         \caption{The outputs of the first two and last two layer in $\mathcal{M}_{32}$ whose posterior $q(\Z)$ is a single Gaussian with the standard deviations initialized from a half normal $\mathcal{N}^+(0.32, 0.02)$.}
         \label{fig:activation_pca_std32}
     \end{subfigure}
    \caption{PCA plots of the outputs for the first two and last two layers on a node-based \textsc{all-cnn-c} BNN with respect to one image from \textsc{cifar10}. Grey unfilled circle are samples from the output distribution induced by the latent variables, while the red ellipse is the $99$ percentile of this distribution. The color circle represents the expected output $\hf^\ell$ under input corruptions, where we fill the circle if it lies within the ellipse. Each axis label is the component index and its explained variance ratio. In the legend, we denote the severity of the corruptions and the ratio between number of points lie within the 99 percentile of the output distribution and the total number of corruption types. We use the corruptions from \cite{hendrycks2019robustness} containing 5 levels of severity and 19 types. For the model with larger $H[q(\Z)]$ in \ref{fig:activation_pca_std32}, the number of points lie within the ellipse is higher than the model with smaller $H[q(\Z)]$ in \ref{fig:activation_pca_std16}.}
    \label{fig:activation_pca}
\end{figure}

\section{Additional details on the experiments and hyperparameters}
\subsection{Approximation for the KL divergence with mixture variational posterior}
We use a mixture of Gaussians (MoG) distribution with $K$ equally-weighted components to provide a flexible approximation of the true posterior in the latent space:
\begin{align}
    q(\Z) = \frac{1}{K} \sum_{k=1}^K q_k(\Z) \\
    q_k(\Z) = \prod_{\ell=1}^L q_{k,\ell}(\Z^\ell) \\
    q_{k,\ell}(\Z^\ell) = \N(\bmu_{k,\ell}, \diag \, \bs_{k,\ell}^2).
\end{align}
where $L$ is the number of layers. We use a Gaussian prior with global scalar variance $s^2$ for the latent prior,
\begin{equation}
    p(\Z) = \N(\1, s^2 I).
\end{equation}

The KL divergence decomposes into cross-entropy and entropy terms,
\begin{align}
    \kl[q(\Z) \, || \, p(\Z)] &= \H[q,p] - \H[q] = \frac{1}{K} \sum_{k=1}^K \H[q_k,p] - \H[q],
\end{align}
where the cross-entropy reduces into tractable terms $\H[q_k,p]$ for Gaussians. The mixture entropy $\H[q]$ remains intractable, but admits a lower bound \cite{entropy},
\begin{align}\label{eq:lb_entropy}
    \H[q] &\ge \frac{1}{K} \sum_{k=1}^K \H[q_k] - \frac{1}{K} \sum_{k=1}^K \log \left( \frac{1}{K} \sum_{r=1}^K \mathrm{BC}(q_k, q_r)\right) \triangleq \widehat{\H}[q]
\end{align}
where
\begin{equation}
    \mathrm{BC}(q,q') = \int \sqrt{q(\z)} \sqrt{q'(\z)} d\z \qquad \le 1
\end{equation}
is the Bhattacharyya kernel of overlap between two distributions \citep{jebara2003bhattacharyya,jebara2004probability}, and has a closed form solution for a pair of Gaussians $q,q'$. The Bhattacharyya kernel has the convenient normalization property $\mathrm{BC}(q,q) = 1$. The lower bound considers unary and pairwise component entropies.

\subsection{Experimental details and hyperparameters}
We actually maximizes the following objective to train the node-based BNNs on large-scale experiments:
\begin{equation}
    \L_{\gamma, \beta}(\htheta, \phi)=\E_{q_{\phi}(\Z)}\Big[\log p(\D | \htheta,\Z)\Big] + \log p(\htheta) + \beta\left(-\H\big[q_{\phi}(\Z), p(\Z)\big] + (\gamma+1) \widehat{\H}\big[q_\phi(\Z)\big] \right)
\end{equation}
which is the augmented ELBO in \cref{eq:entropy_elbo} with additional coefficient $\beta$ for the cross-entropy and variational entropy term.
We also replace the intractable mixture entropy $\H[q]$ with its tractable lower bound $\widehat{\H}[q]$ presented in \cref{eq:lb_entropy}.
During training, we will anneal $\beta$ from $0$ to $1$. We found this to have ease optimization and produce better final results. For all experiments, we estimate the expected log-likelihood in the loss function using 4 samples.

For all the experiments on \textsc{cifar10}/\textsc{cifar100}, we run each experiment for 300 epochs, where we increase $\beta$ from $0$ to $1$ for the first 200 epochs. We use SGD as our optimizer, and we use a weight decay of $0.0005$ for the parameters $\theta$. We use a batch size of 128.
For all the experiments on \textsc{tinyimagenet}, we run each experiment for 150 epochs, where we increase $\beta$ from $0$ to $1$ for the first 100 epochs. We use a batch size of 256.
Bellow, we use $\lambda_1$ and $\lambda_2$ to denote the learning rate of the parameters $\theta$ and $\phi$ respectively.

For \textsc{vgg16}, we set the initial learning rate $\lambda_1=\lambda_2=0.05$, and we decrease $\lambda_1$ linearly from $0.05$ to $0.0005$ from epoch 150 to epoch 270, while keeping $\lambda_2$ fixed throughout training. We
initialize the standard deviations with $\mathcal{N}^+(0.30, 0.02)$ and set the standard deviation of the prior to $0.30$.

For \textsc{resnet18}, we set the initial learning rate $\lambda_1=\lambda_2=0.10$, and we decrease $\lambda_1$ linearly from $0.10$ to $0.001$ from epoch 150 to epoch 270, while keeping $\lambda_2$ fixed throughout training. We
initialize the standard deviations with $\mathcal{N}^+(0.40, 0.02)$ and set the standard deviation of the prior to $0.40$.

For \textsc{preactresnet18}, we set the initial learning rate $\lambda_1=\lambda_2=0.10$, and we decrease $\lambda_1$ linearly from $0.10$ to $0.001$ from epoch 75 to epoch 135, while keeping $\lambda_2$ fixed throughout training. We
initialize the standard deviations with $\mathcal{N}^+(0.30, 0.02)$ and set the standard deviation of the prior to $0.30$.

\subsection{Runtime}
We report the average running times of different methods in Table \ref{tab:runtime}.
We used similar number of epochs for all methods in each experiment.
All experiment were performed on one Tesla V100 GPU.
Overall, node BNNs took 4 times longer to train than SWAG since we use 4 Monte Carlo samples per training sample to estimate the expected log-likelihood in the $\gamma$-ELBO.
ASAM took 2 times longer to train than SWAG since they require two forward-backward passes per minibatch.

\begin{table}[th]
\centering
\caption{Average running times of different methods measured in seconds. All experiments were performed on one Tesla V100 GPU.}
\label{tab:runtime}
\begin{tabular}{@{}lllll@{}}
\toprule
Model          & Dataset      & Node-BNN & SWAG  & ASAM  \\ \midrule
VGG16          & CIFAR100     & 13274    & 3384  & 6870  \\
               & CIFAR10      & 12941    & 3251  & 6539  \\ \midrule
ResNet18       & CIFAR100     & 18093    & 4528  & 9086  \\
               & CIFAR10      & 17733    & 4474  & 8921  \\ \midrule
PreActResNet18 & TinyImagenet & 54892    & 13830 & 26564 \\ \bottomrule
\end{tabular}
\end{table}

\section{Additional benchmark results}\label{app:benchmark}
\begin{figure}[t]
    \centering
    \includegraphics[width=.7\columnwidth]{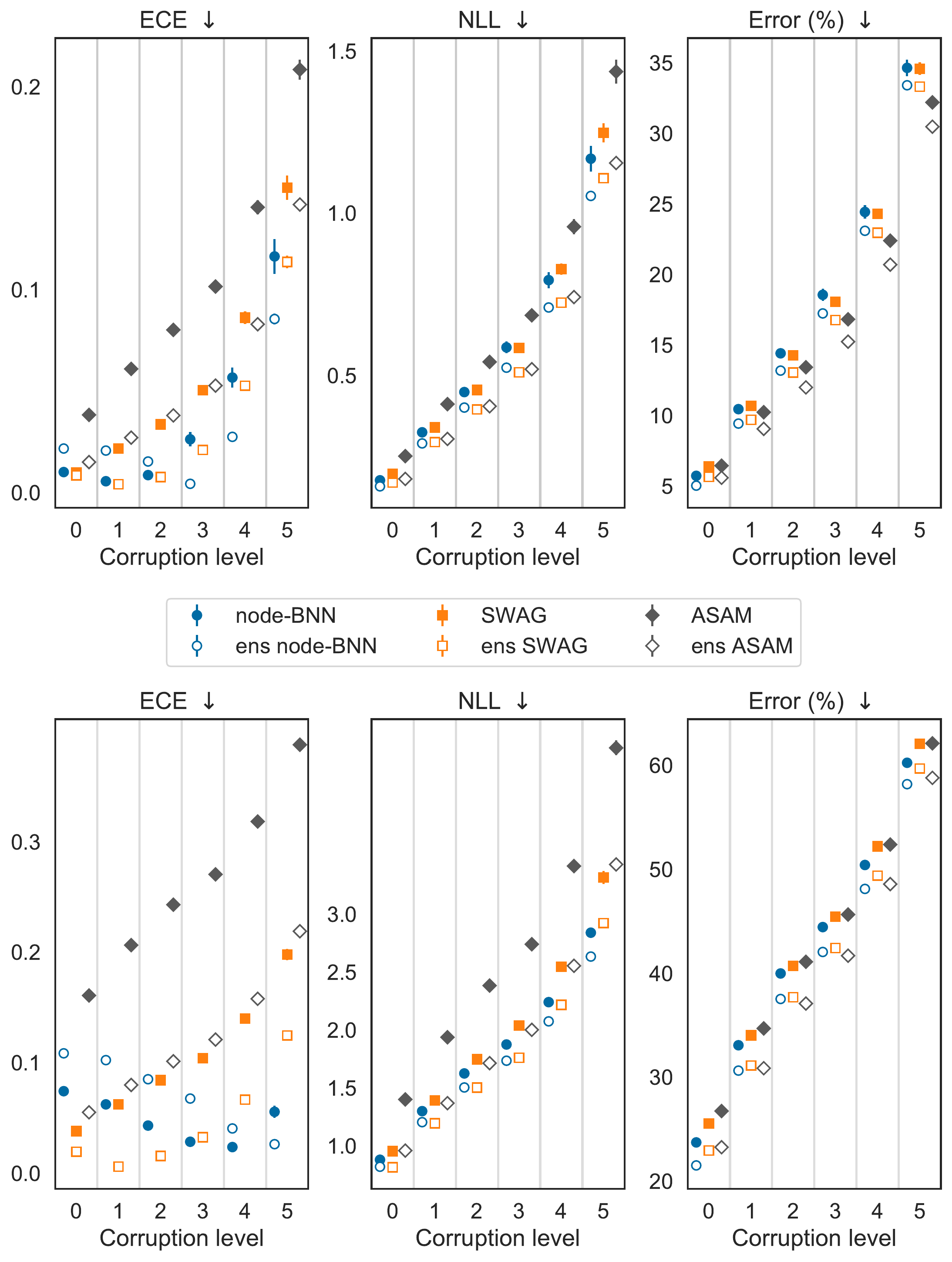}
    \caption{Results of \textsc{vgg16} on \textsc{cifar10} \textbf{(top)} and \textsc{cifar100} \textbf{(bottom)}. We use $K=4$ and only the latent output variables for node-based BNNs. We plot ECE, NLL and error for different corruption levels, where level $0$ indicates no corruption. We report the average performance over 19 corruption types for level 1 to 5. We denote the ensemble of a method using the shorthand \emph{ens} in front of the name. Each result is the average over 25 runs for non-ens versions and 5 runs for ens versions. The error bars represent the standard deviations across different runs. Node-based BNNs and their ensembles (\textcolor[RGB]{0,107,164}{blue}) perform best in term of ECE and NLL on OOD data, while having similar accuracy to other methods.}
    \label{fig:benchmark_vgg16}
\end{figure}

\begin{figure}[t]
    \centering
    \includegraphics[width=.8\columnwidth]{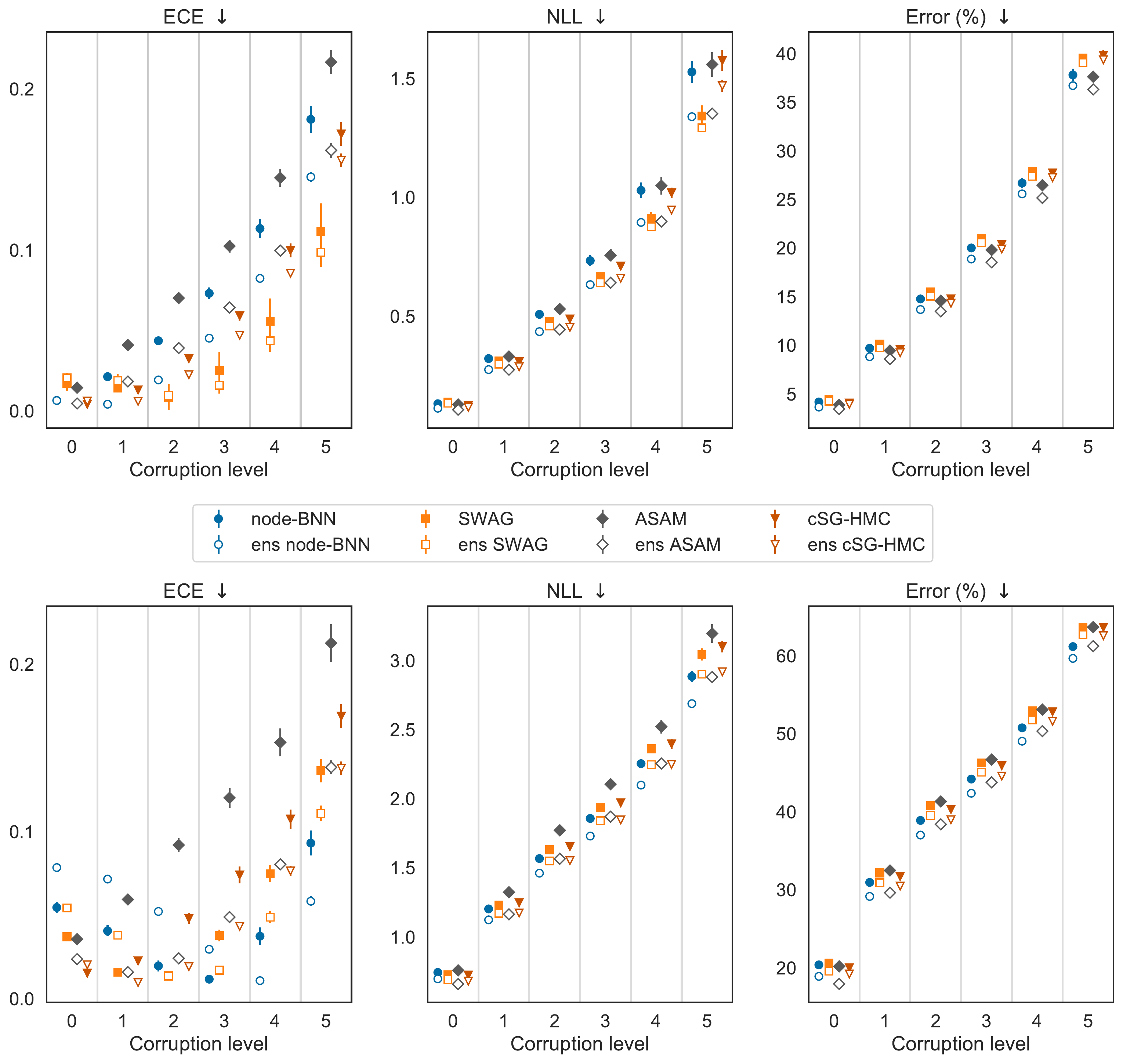}
    \caption{Results of \textsc{resnet18} on \textsc{cifar10} \textbf{(top)} and \textsc{cifar100} \textbf{(bottom)}. We use $K=4$ and only the latent output variables for node-based BNNs. We plot ECE, NLL and error for different corruption levels, where level $0$ indicates no corruption. We report the average performance over 19 corruption types for level 1 to 5. We denote the ensemble of a method using the shorthand \emph{ens} in front of the name. Each result is the average over 25 runs for \emph{non-ens} versions and 5 runs for \emph{ens} versions. The error bars represent the standard deviations across different runs. Node-based BNNs and their ensembles (\textcolor[RGB]{0,107,164}{blue}) perform best across all metrics on OOD data of \textsc{cifar100}, while having competitive results on \textsc{cifar10}.}
    \label{fig:benchmark_resnet18_appendix}
\end{figure}

\begin{figure}[t]
    \centering
    \includegraphics[width=.8\columnwidth]{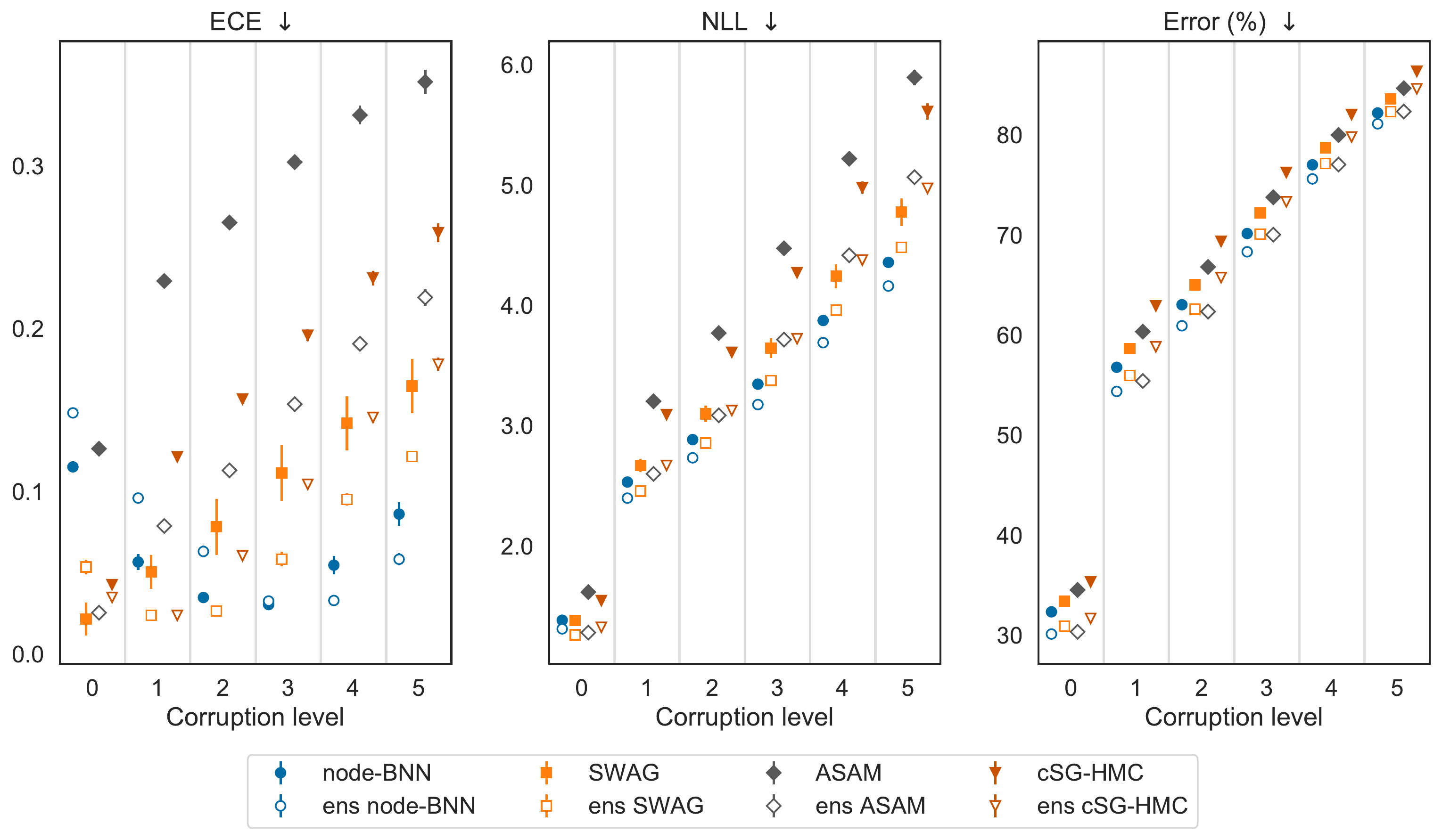}
    \caption{Results of \textsc{preactresnet18} on \textsc{tinyimagenet}. We use $K=4$ and only the latent output variables for node-based BNNs. We plot ECE, NLL and error for different corruption levels, where level $0$ indicates no corruption. We report the average performance over 19 corruption types for level 1 to 5. We denote the ensemble of a method using the shorthand \emph{ens} in front of the name. Each result is the average over 25 runs for \emph{non-ens} versions and 5 runs for \emph{ens} versions. The error bars represent the standard deviations across different runs. Node-based BNNs and their ensembles (\textcolor[RGB]{0,107,164}{blue}) perform best accross all metrics on OOD data, while having competitive performance on ID data.}
    \label{fig:benchmark_preactresnet18_appendix}
\end{figure}

Here we include the benchmark results of \textsc{vgg16} on \textsc{cifar10} and \textsc{cifar100} in \cref{fig:benchmark_vgg16}. We also include \cref{fig:benchmark_resnet18_appendix} and \cref{fig:benchmark_preactresnet18_appendix} as larger versions of \cref{fig:benchmark_resnet18} and \cref{fig:benchmark_preeacresnet18}.
\section{The evolution of variational entropy during training}\label{sec:entropy_evolution}
\begin{figure}[t]
    \centering
    \includegraphics[width=\textwidth]{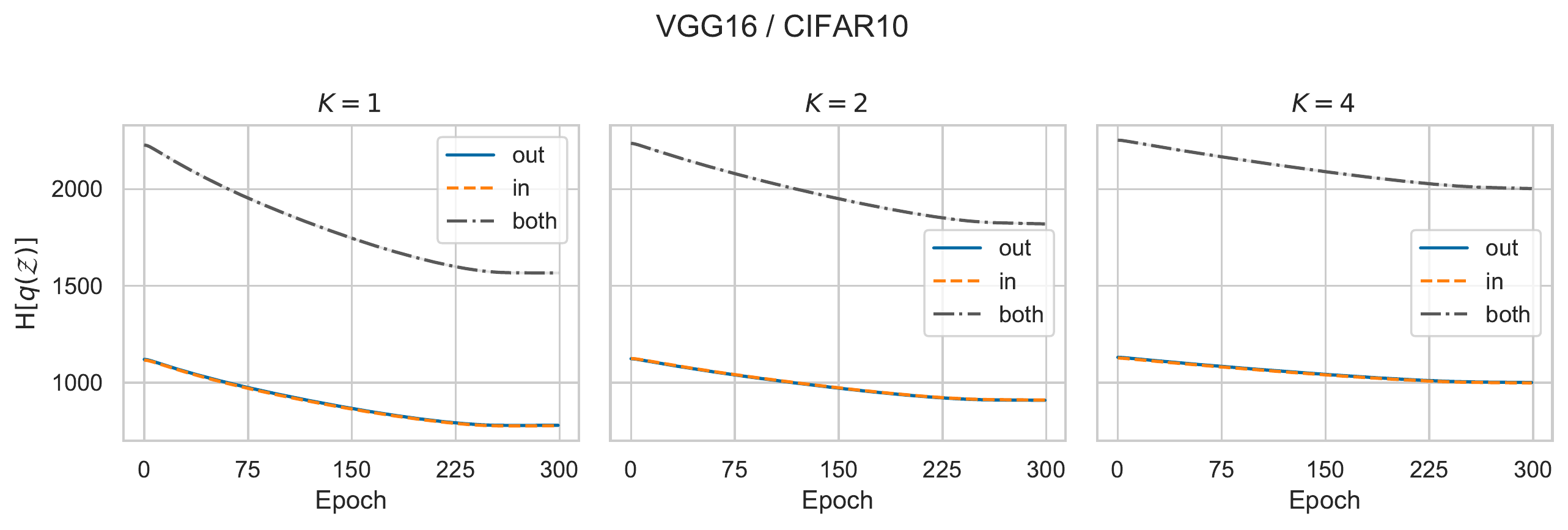}
    \caption{The evolution of entropy during training for \textsc{vgg16} / \textsc{cifar10} when trained using the original ELBO. Each result is averaged over 5 runs. Each error bar represents one standard deviation but it is too small to be seen.}
    \label{fig:vgg16_cifar10_entropy}
\end{figure}
\begin{figure}[t]
    \centering
    \includegraphics[width=\textwidth]{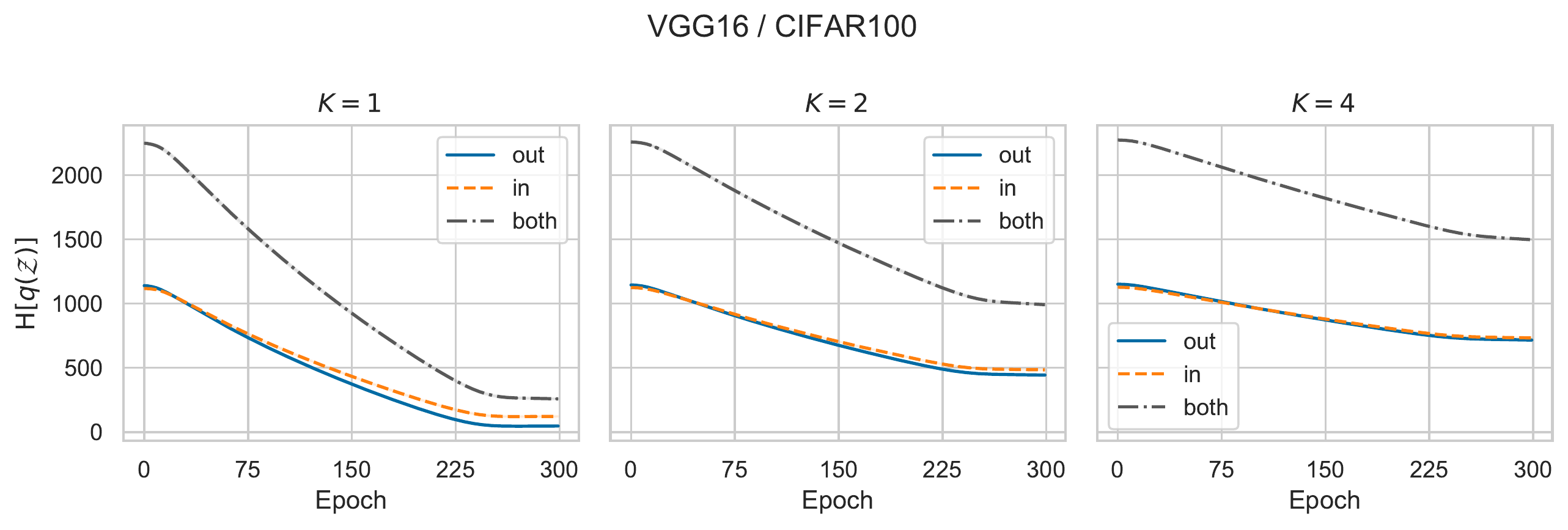}
    \caption{The evolution of entropy during training for \textsc{vgg16} / \textsc{cifar100} when trained using the original ELBO. Each result is averaged over 5 runs. Each error bar represents one standard deviation but it is too small to be seen.}
    \label{fig:vgg16_cifar100_entropy}
\end{figure}
\begin{figure}[t]
    \centering
    \includegraphics[width=\textwidth]{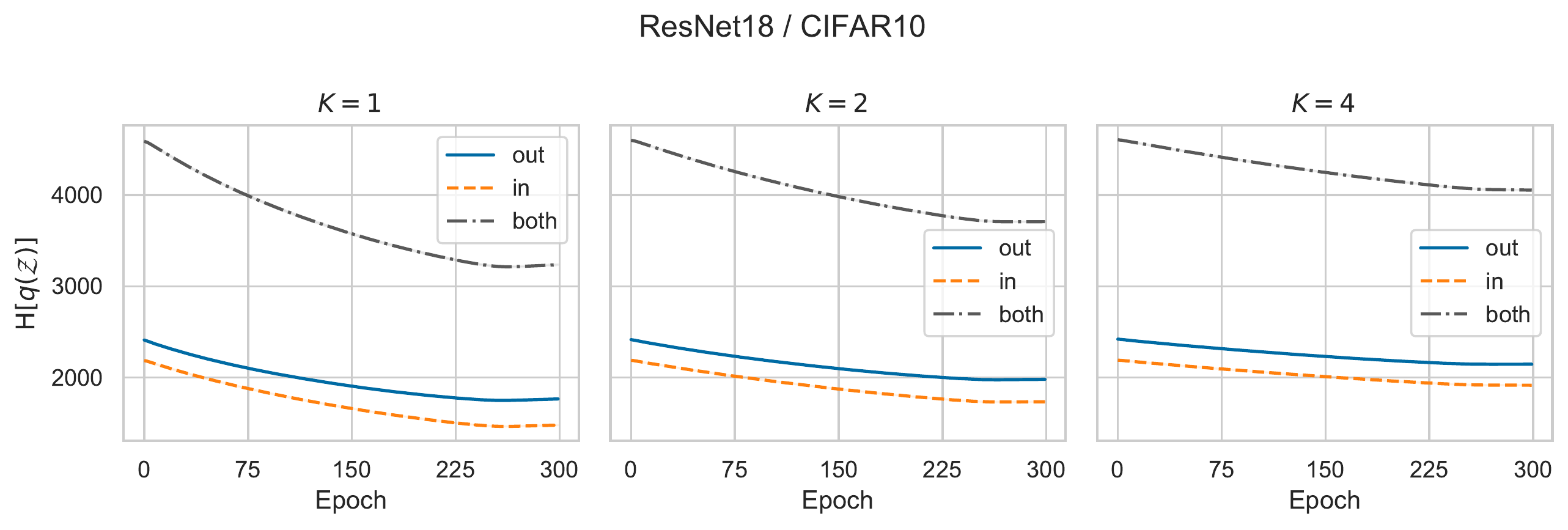}
    \caption{The evolution of entropy during training for \textsc{resnet18} / \textsc{cifar10} when trained using the original ELBO. Each result is averaged over 5 runs. Each error bar represents one standard deviation but it is too small to be seen.}
    \label{fig:resnet18_cifar10_entropy}
\end{figure}
\begin{figure}[t]
    \centering
    \includegraphics[width=\textwidth]{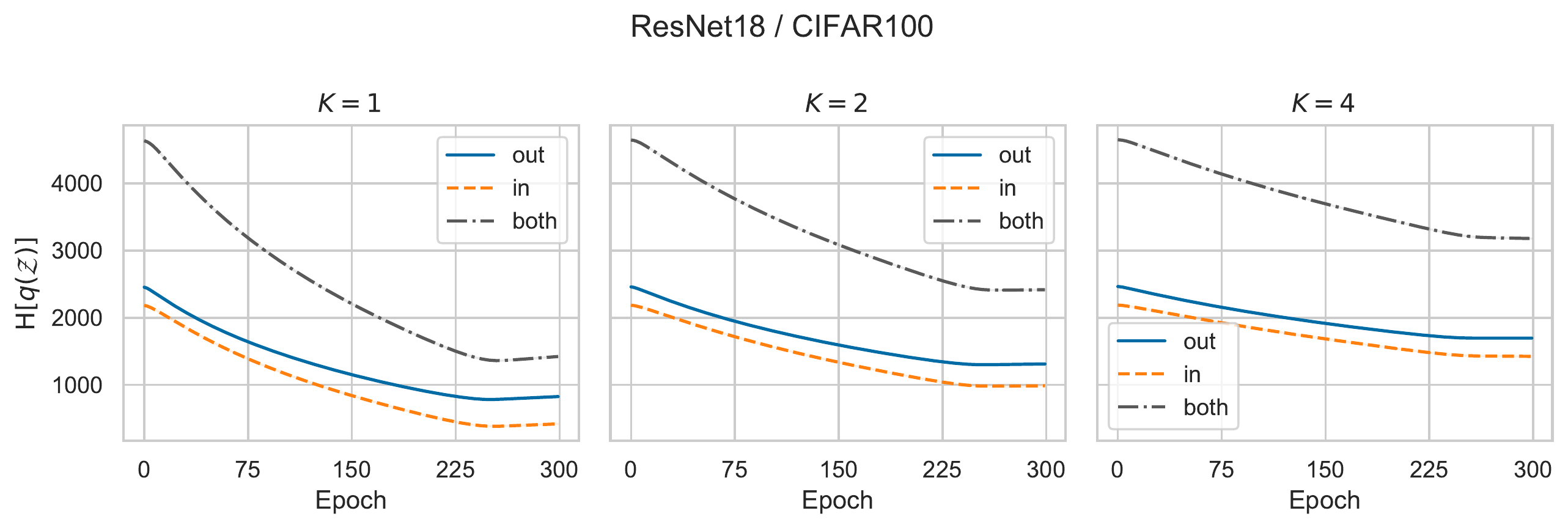}
    \caption{The evolution of entropy during training for \textsc{resnet18} / \textsc{cifar100} when trained using the original ELBO. Each result is averaged over 5 runs. Each error bar represents one standard deviation but it is too small to be seen.}
    \label{fig:resnet18_cifar100_entropy}
\end{figure}
We visualize the progression of the variational entropy when trained using the original ELBO (without the $\gamma$-entropy term) under different settings in \cref{fig:vgg16_cifar10_entropy,fig:vgg16_cifar100_entropy,fig:resnet18_cifar10_entropy,fig:resnet18_cifar100_entropy}.
We can observe the typical behaviour of variational inference that it tends to reduce the entropy of the variational posterior over time.

\section{Additional results on the effect of $\gamma$ on performance of node-based BNNs}\label{app:gamma_effect}
\begin{figure}[ht]
     \centering
     \includegraphics[width=\textwidth]{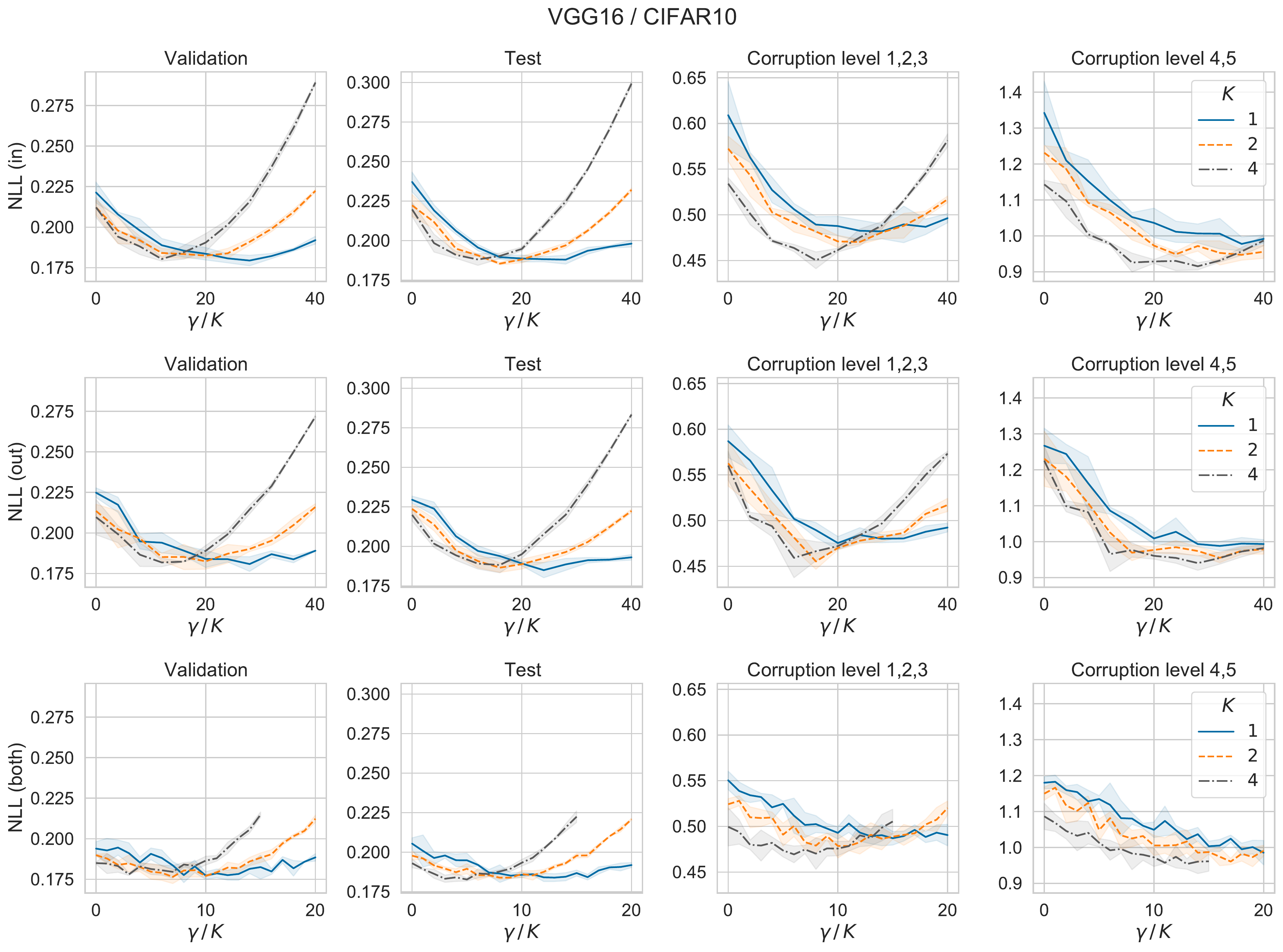}
     \caption{Results of \textsc{vgg16} on \textsc{cifar10} under different $\gamma$ value. $K$ is the number of components.
     Each row corresponds a different latent variable structure. We report the mean and standard deviation over 5 runs for each result.}
     \label{fig:stovgg16_cifar10_nll}
 \end{figure}

\begin{figure}[ht]
     \centering
     \includegraphics[width=\textwidth]{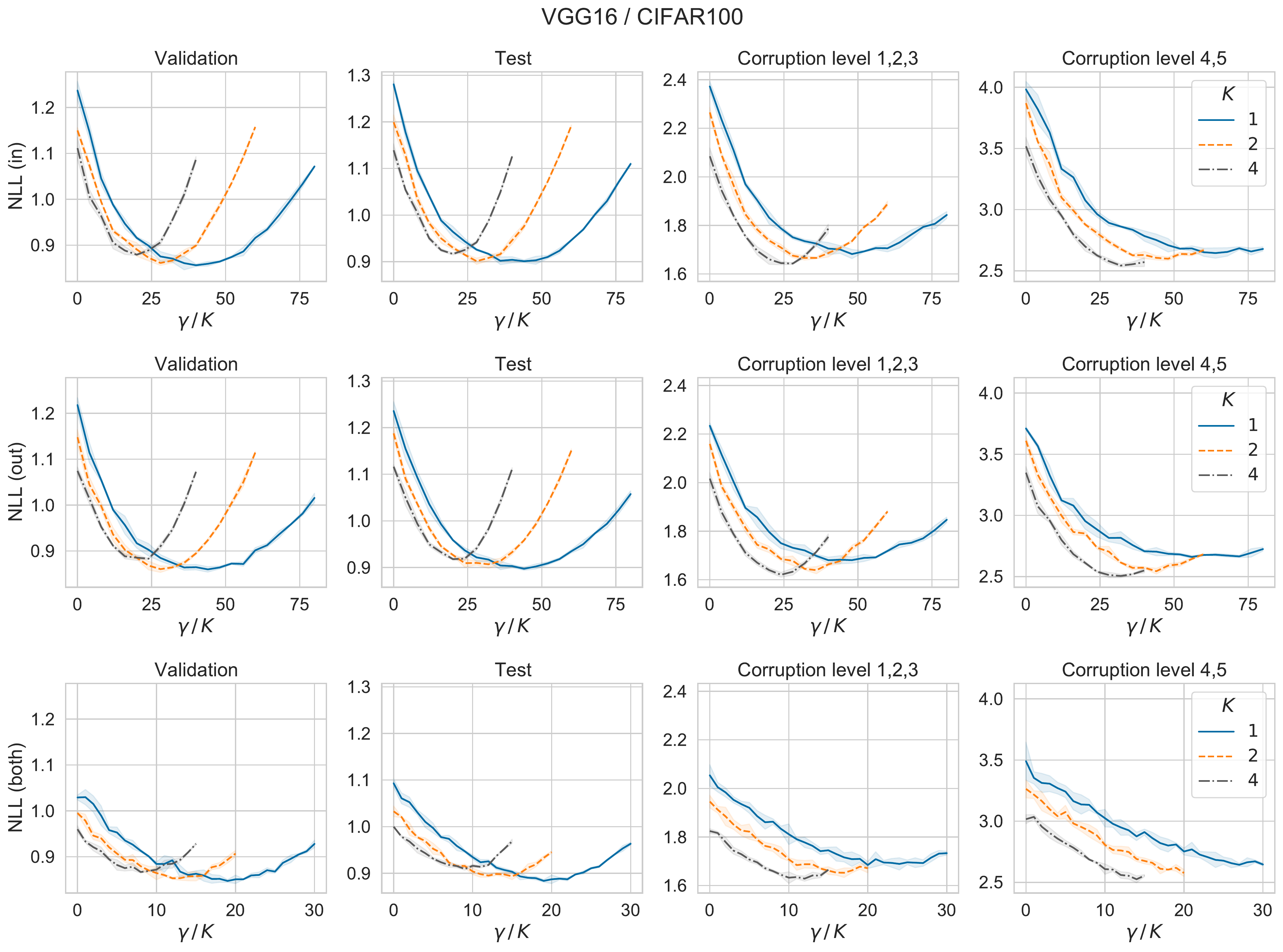}
     \caption{Results of \textsc{vgg16} on \textsc{cifar100} under different $\gamma$ value. $K$ is the number of components.
     Each row corresponds a different latent variable structure. We report the mean and standard deviation over 5 runs for each result.}
     \label{fig:stovgg16_cifar100_nll}
 \end{figure}

\begin{figure}[ht]
     \centering
     \includegraphics[width=\textwidth]{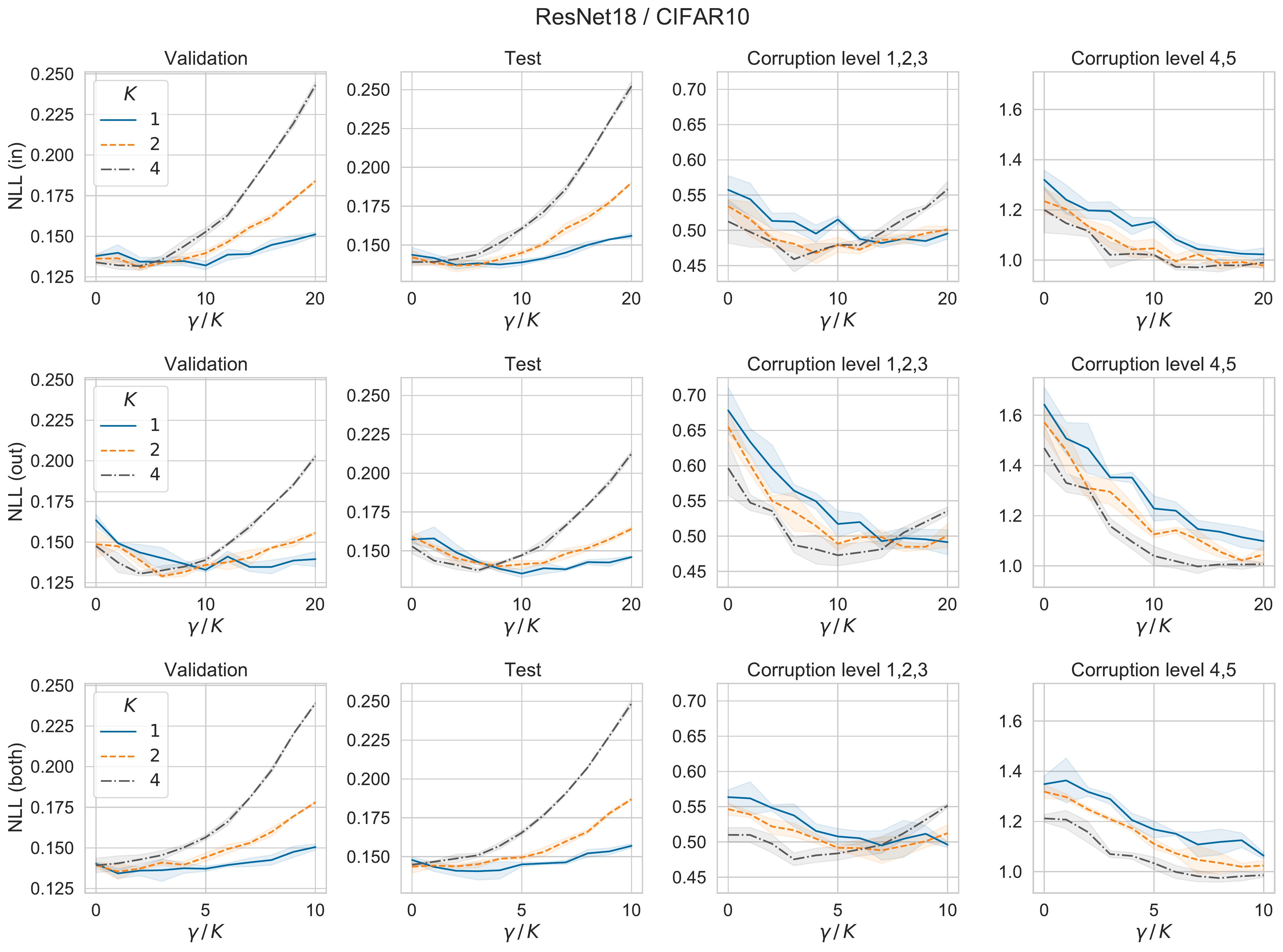}
     \caption{Results of \textsc{resnet18} on \textsc{cifar10} under different $\gamma$ value. $K$ is the number of components. Each row corresponds a different latent variable structure. We report the mean and standard deviation over 5 runs for each result.}
     \label{fig:storesnet18_cifar10_nll}
 \end{figure}

\begin{figure}[ht]
     \centering
     \includegraphics[width=\textwidth]{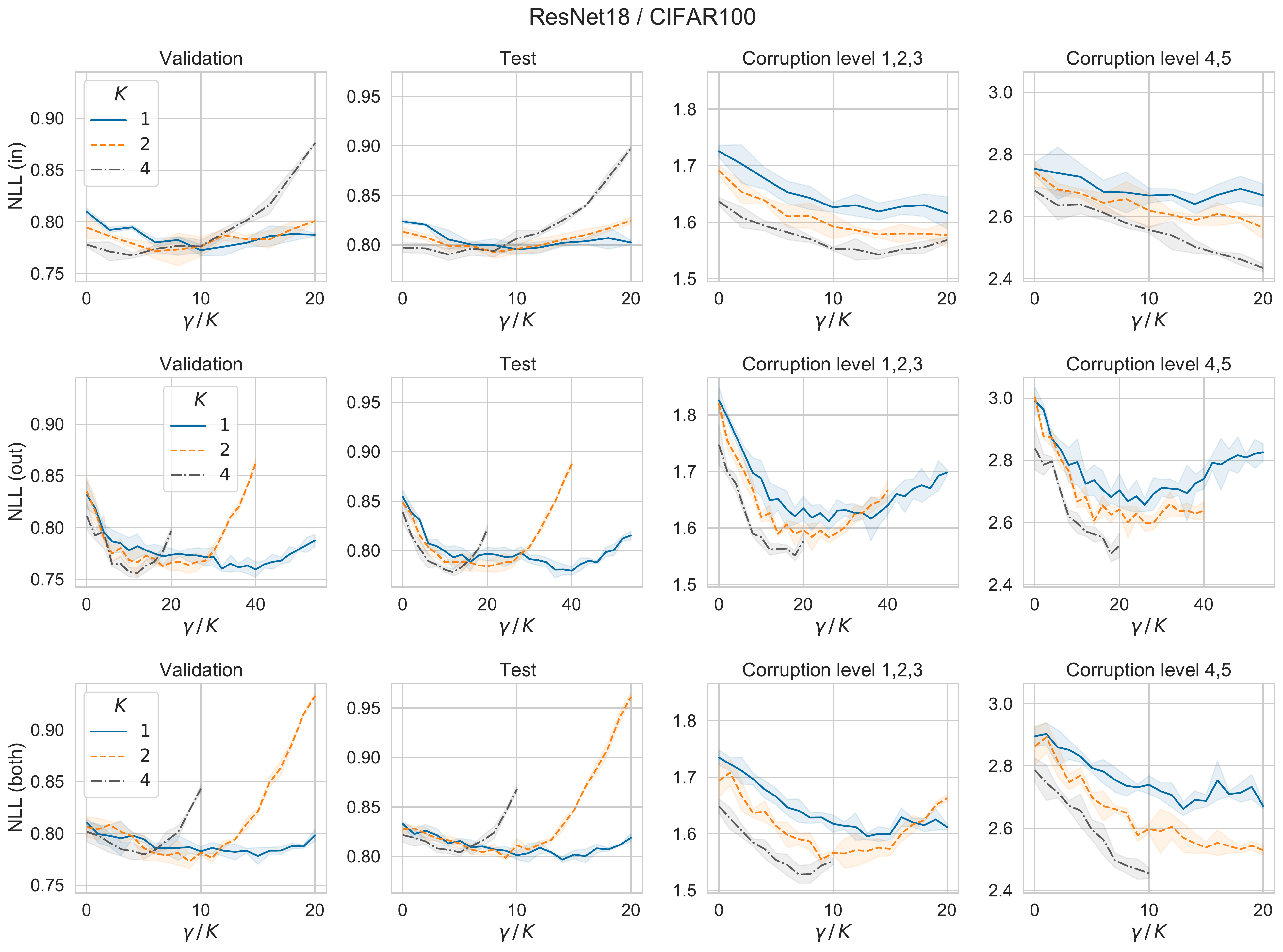}
     \caption{Results of \textsc{resnet18} on \textsc{cifar100} under different $\gamma$ value. $K$ is the number of components. Each row corresponds a different latent variable structure. We report the mean and standard deviation over 5 runs for each result.}
     \label{fig:storesnet18_cifar100_nll}
 \end{figure}

\begin{figure}[ht]
     \centering
     \includegraphics[width=\textwidth]{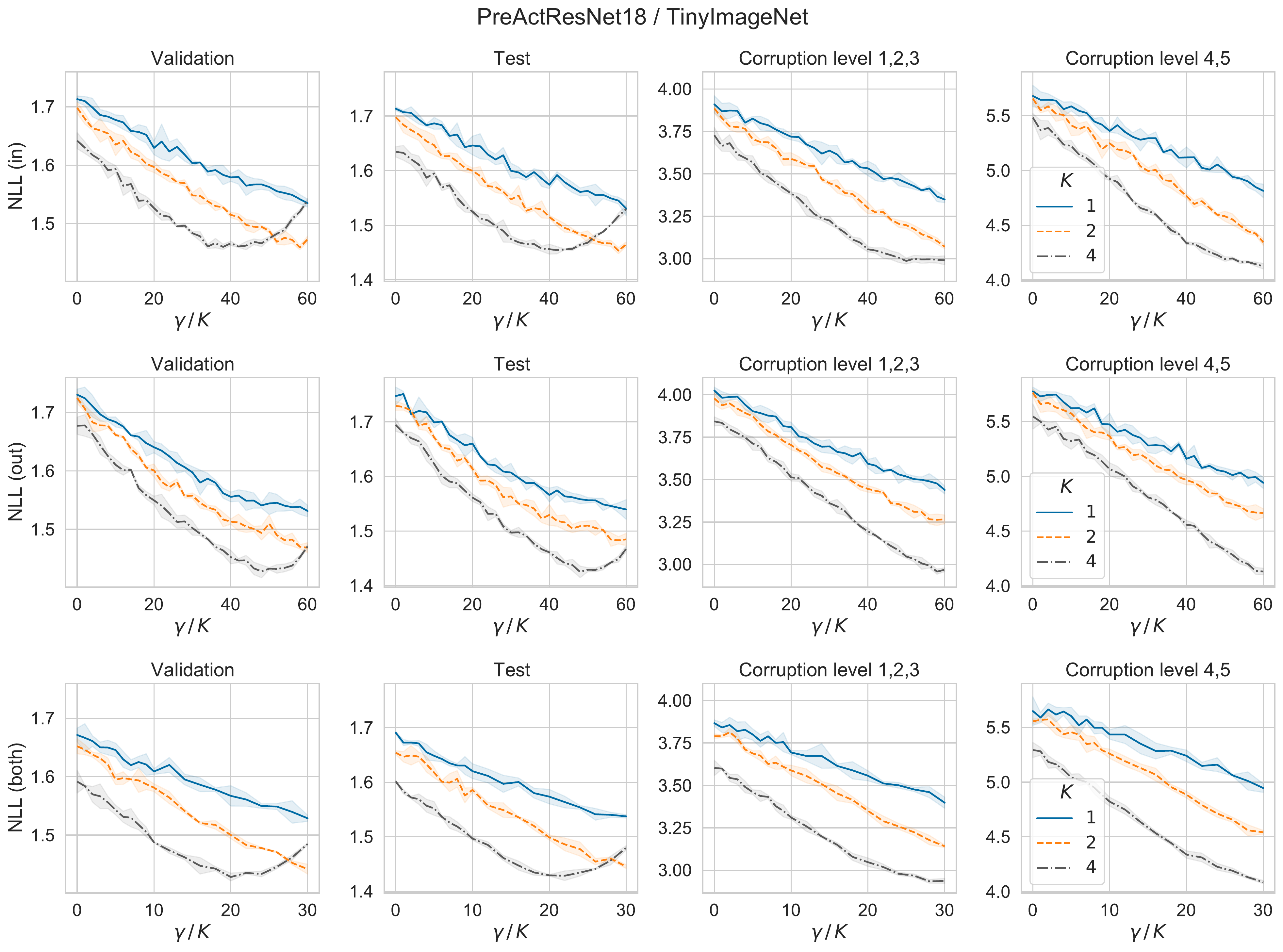}
     \caption{Results of \textsc{preactresnet18} on \textsc{tinyimagenet} under different $\gamma$ value. $K$ is the number of components. Each row corresponds a different latent variable structure. We report the mean and standard deviation over 5 runs for each result.}
     \label{fig:stopreactresnet18_tinyimagenet_nll}
 \end{figure}

Here we include \cref{fig:stovgg16_cifar10_nll,fig:stovgg16_cifar100_nll,fig:storesnet18_cifar10_nll,fig:storesnet18_cifar100_nll,fig:stopreactresnet18_tinyimagenet_nll} to show the effect of $\gamma$ on performance of node-based BNNs under different architectures and datasets.


\end{document}